# A Framework for Characterizing Novel Environment Transformations in General Environments

Matthew Molineaux[1], Dustin Dannenhauer[1], Eric Kildebeck[2]

[1]Parallax Advanced Research, Beavercreek, OH 45431
*firstname.lastname@parallaxresearch.org*

[2]University of Texas at Dallas, Richardson, TX 75080
*eric.kildebeck@utdallas.edu*

## Abstract

To be robust to surprising developments, an intelligent agent must be able to respond to many different types of unexpected change in the world. To date, there are no general frameworks for defining and characterizing the types of environment changes that are possible. We introduce a formal and theoretical framework for defining and categorizing environment transformations, changes to the world an agent inhabits. We introduce two types of environment transformation: R-transformations which modify environment dynamics and T-transformations which modify the generation process that produces scenarios. We present a new language for describing domains, scenario generators, and transformations, called the Transformation and Simulator Abstraction Language (T-SAL), and a logical formalism that rigorously defines these concepts.  Then, we offer the first formal and computational set of tests for eight categories of environment transformations. This domain-independent framework paves the way for describing unambiguous classes of novelty, constrained and domain-independent random generation of environment transformations, replication of environment transformation studies, and fair evaluation of  agent robustness.

## 1    Introduction

A primary and unrealized goal of artificial intelligence is *robustness*: the capability to continue to perform well despite novel changes in environments an agent interacts with. Robustness is pivotal to continued autonomy in open worlds – those that are dynamic and unstable like the physical world. To date, the science of machine learning has been assumed to prepare agents to become more robust without rigorously defining what the space of such novel changes are, nor how they can be systematically varied to test the robustness of an agent. Descriptions of concept learning or classification learning problems in machine learning (Mitchell, 1997) traditionally give a set of examples from a fixed distribution, with no described temporal relationships. Reinforcement learning problems (Sutton and Barto, 2018) describe an environment that an agent receives observations from, but generally assume the environment is unchanging. Recent work on open-world novelty (Boult et al, 2021; Muhammad et al, 2021; Gamage et al, 2021), characterizes changes to an environment from the point of view of the agent, preventing direct comparisons across agents with different world views. Langley (2020) describes environmental change, but without a rigorous framework.

To better describe the challenges that a changing environment poses to an agent, we define a general class of environments and *environment transformations* (sometimes referred to as *novelties*). These definitions are useful for 1) describing the extent of environment transformations, 2) categorizing and grouping transformations with similar characteristics, 3) communicating and replicating transformations, and 4) randomly generating novelties that test the robustness of an agent to novel changes in its environment. To our knowledge, this is the first framework of definitions that describes formally and completely a space of environment transformations over a general class of environments.

Transformations to environments come in two general categories: *R-Transformations* to the dynamics of the environment, and *T-Transformations* to scenario distributions ; the latter category includes distributions over possible initial states and performance criteria. To rigorously define a general class of environments and novelties, we start with a well-known and tested formalism for representing these environments, the planning domain definition language (PDDL) family of languages. PDDL provides a general mechanism for describing individual scenarios referred to as "problems", and a separate mechanism for describing environment dynamics referred to as "domains". In this work, we formally describe the general notion of a "scenario generator", which describes a distribution of scenarios rather than a single scenario.

T-SAL R-Transformations describe changes to domains as used in PDDL. PDDL domains use a subset of first-order logic to describe environment dynamics: changes that may and must happen as a result of agent actions and the natural interactions of objects without agency. Using established PDDL formalisms, we build a new language, the T-SAL (Transformation and Simulator Abstraction Language) Domain language, that covers the broadest class of environments possible. Unlike prior work on PDDL languages, we do not constrain the representation of T-SAL based on backwards compatibility or to match well the capabilities of existing automated planners; therefore, we incorporate events and processes from PDDL+ that describe continuous change, non-deterministic effects and events that describe change probabilistically, mathematical operators that describe proportional change, and an object introduction effect that describes open-world change.

T-SAL T-Transformations describe changes to scenario generators. In addition to the initial state classically described within a PDDL problem, these scenario generators describe how to draw initial state objects, and fluents from statistical distributions. Instead of a goal, scenario generators include a performance function. This is similar to existing maximization goals, but defined with respect to agents as well as the state, permitting multiple agents' performance to be described.

T-Transformations and R-Transformations are useful both to generate and to formally describe novelty. This work contributes to the literature on open-world novelty by formally characterizing a group of novelty concepts previously described only informally. Scientists working together in the DARPA SAIL-ON program to organize and schedule program advancement developed a group of characterizations described as a "novelty hierarchy". This novelty grouping was previously defined only informally, so some practitioners may reasonably disagree with our definitions; however, we believe they capture the essence of the designers' intent. By providing this formal definition of the novelty hierarchy used in early novelty investigations, we hope to show that T-SAL transformations are a viable means of rigorously categorizing novelty in the future. We believe this forms a suitable basis for future investigations of the impact of novelty categories on agents.

In the following sections, we (1) formally define T-SAL domains, (2) give formulas for assessing T-SAL domain legality, (3) formally describe scenario generators, (4) formally describe transformations over T-SAL domains, (5), formally describe transformations over scenario generators, and (6) formally describe the SAIL-ON program novelty levels. A formalization for the state transition function of T-SAL is not given, but is based on PDDL+ (Fox and Long 2006) with only small modifications necessary. See Appendix C for a list of the differences between T-SAL and PDDL+.

## 2 Related Work

In this section, we review prior work that informed the development of our framework. In the literature on creativity, novelty is a well-defined concept. Wiggins' (2006) Creative Systems Framework (CSF), which formally defined and extended the creativity system described by Boden (1990) in her book *The Creative Mind*, defines novelty as a property of a creative output which previously did not exist. CSF was the first formal logic-based creativity framework, describing, at the highest level, a conceptual space of universes and artifacts that naturalistic, creative agents create in a societal context. Our framework is both narrower in scope and provides greater detail, ranging over transformations to environments that consist of agents, actions, events, processes, and tasks, commonly found in AI literature. In Wiggins' CSF, $\mathcal{R}$ denotes a set of universe constraining rules, and $\mathcal{T}$ denotes rules for traversing a space. We maintain consistency with Wiggins' prior definitions of $\mathcal{R}$ and $\mathcal{T}$, dividing our transformations into R-Transformations and T-Transformations respectively.

The term *novelty* in AI literature has taken on at least two perspectives: novelty from the lens of an individual agent and novelty that exists from the lens of an environments' history. Research of the former defines novelty relative to an agent's experiences (e.g., Muhammad et al. (2021), Boult et al. (2021), Gamage et al., (2021)), while our framework defines novelty as the latter, consistent with work on creativity (Wiggins', 2006). Frameworks with an agent's perspective of novelty are useful because they can identify how prepared an agent is for new challenges. However, they cannot usefully describe how environment challenges differ in a way that cuts across different agents and different knowledge representations. This limits the ability for agent-based notions of novelty to support evaluations on a variety of different AI approaches. For historical and language reasons, both areas of research currently use the term "novelty", but in different ways; therefore we emphasize our novelty framework is environment-based.

Environment-based novelty is fundamentally different from agent-based novelty. In the theory of open world novelty by Boult et al. (2021), novelty occurs when an agent experiences an environment sufficiently different from its prior experiences. A key feature being that novelty occurs at a point in time. In environment-based frameworks like that of Wiggins' CSF (2006), novelty exists independent of any given agent's experiences or knowledge. Additionally, our framework does not define novelty as occurring at a point in time, but rather as encompassing the time-independent difference between two environments. This means that novelty can exist without any changes to the observation or state space. While valuable for considering what experiences could be new to a particular agent, two agents cannot truly be compared on the "same" novelty unless they have identical prior experiences; such comparisons are a key feature of our framework.

A need for a theory of environmental change has been proposed to help explain and measure progress in open-world learning. Langley's (2020) work provides a set of broad requirements for such a theory, motivating a theoretical formalism for environments and transformations on them. Our framework provides both, and so can be consider an example "theory" in this regard that describes environmental changes and provides specific language for characterizing how environment changes vary.

Existing novelty generation capabilities for evaluating AI systems are designed in close alignment with simulated environments. Current benchmark domains (Goel et al. (2021), Xue et al., (2022), Kejriwal & Thomas (2021), Balloch et al., (2022)) test an agent's ability to respond to novelty using carefully constructed scenarios by domain experts. This is current standard practice, and each class of transformations constructed applies only to a single environment. This ad-hoc construction does not permit generalization over domains or formal statements about what characteristics apply to the novel scenarios generated. To our best knowledge, our framework is the first to systematically provide infinite novelties that are not hand curated by domain experts.

Game design research has explored problems of domain generation and notions of transformations. The Metagame system (Pell, 1992) and EGGG system (Orwant, 2000) automatically generated chess-like games using a complicated system of rules and constraints designed for chess-like games in particular. Smith and Mateas (2010) decoupled the design of a game space from the exploration of that space to find interesting games. The language used to design game spaces was large and flexible compared to the prior chess-like games, but still incorporated many assumptions, such as the existence of a rectangular play space. Our framework's language for expressing domains, based on PDDL, can model many aspects found in the game's representations of these works; the aim of our framework is to produce novelty first and then meet other criteria later (such as being an interesting game). Therefore we focus on a more fundamental representation without worrying about domain specific knowledge beyond representational notions of objects, actions, events, and goals. To sculpt the generative space of our framework, users would provide constraints on the choice of R and T-transformations as they see fit – and such a discussion is outside the scope of this paper.

Prior work (Molineaux & Dannenhauer, 2021) defined a set of environment transformation characteristics, or "dimensions", useful for classifying the difference between pairs of original and modified environments. These definitions are based on describing environments using a refinement of the Partially Observable Markov Decision Process formalism, rather than the relational manipulations described here. The dimension-based formalism is concerned primarily with broad characterizations of the start and endpoints rather than describing a specific pathway between them. As such, it is more difficult to verify, and does not account for human-intuitive descriptions of novelty categories such as those found in the novelty hierarchy. However, the upside there is much less flexibility in the dimension-based formalism to describe the same environments and transformations in different ways; as such, dimension values have much less "wiggle room" than an arbitrary T-SAL categorization.

## 3    Formalism

T-SAL is a general term that refers to two languages for describing environments: the computational representation (T-SAL-CR), and the logical language that describes it, L(T-SAL), *read as 'logic of T-SAL'*. L(T-SAL) is a first-order logic that describes domains, scenario generators, and environment transformations at an abstract level. It provides common terminology and is used for rigorous definitions. T-SAL-CR describes individual domains, using domain-specific fluents and types to describe the dynamics of domains. We build up both in this formalism; L(T-SAL) is used to define and make general statements about environments and environment transformations. The T-SAL-CR is provided to communicate examples. An exact semantics for the T-SAL-CR is not provided here; however, it's similar to existing computational representations in the PDDL family (see Appendix C for a detailed comparison). All examples of T-SAL concepts use the well-understood Cart-Pole domain, supplemented with moving blocks and multiple carts.

## 3.1    T-SAL Domains

T-SAL domains describe properties and dynamics of environments. This section builds up the concept of a domain by describing simpler concepts used in the description of a domain in a bottom-up fashion. A BNF for the T-SAL domain language is given in Appendix D. All variables introduced throughout the rest of this section are shown in Table 1. Short definitions and examples of major T-SAL domain concepts are given in Table 2.

| | |
|---|---|
| *S:* the set of symbols | Alt: the set of effect modifications {SET, INCREASE, DECREASE, CREATE} |
| *V:* variables in a domain description | |
| ***V***: the space of variable sets | Dir: the set of change directions {INCREASE, DECREASE} |
| *Te:* terms which may be of the form of symbols, variables, integers, or reals, | Ev: the space of events |
| *ty*: a type in the space of symbols S | **Ev**: the space of sets of events |
| | ev: an event in the space of events Ev |
| F: the space of functions | P: the space of process models |
| f: a function in the space of functions F | **P**: the space of sets of process models |
| Fn: the set of function names | p: a process model in the space of process models |
| fn: a function name | D: the space of T-SAL domains |
| Cn: the space of conditions | d: a domain in the space of T-SAL domains D |
| **Cn**: the space of sequences of conditions | st: a T-SAL state that is true in the environment at a particular moment in time |
| cnd: a condition in the space of conditions Cn | og: an object generator |
| Ca: the space of calculations | OG: the space of object generators |
| **Ca**: the space of sets of calculations | vg: a value generator |
| calc: a calculation in the space of calculations Ca | VG: the space of value generators |
| Ax: the space of axioms | **Va**: the space of values |
| **Ax**: the space of sets of axioms | fg: a fluent generator |
| ax: an axiom in the space of axioms Ax | sg: a T-SAL scenario generator |
| E: the space of effects | SG: the space of scenario generators |
| **E**: the space of sets of effects | GF: the space of ground fluents |
| e: an effect in the space of effects E | gf: a ground fluent in the space of ground fluents |
| C: the space of continuous changes | AG: the space of agents |
| **C**: the space of sets of continuous changes | ag: an agent in the space of agents |
| c: a continuous change in the space of continuous changes C | t: an environment transformation |
| | **t**: a sequence of environment transformations |
| A: the space of action models | T: the space of transformation sequences |
| **A**: the space of sets of actions models | U: the universal set |
| a: an action model in the space of action models A | |

Table 1 – Summary of variables used in L(T-SAL)

Both L(T-SAL) and  T-SAL-CR borrow heavily from the PDDL (McDermott, et al. (1998); Fox & Long, (2003); Edelkamp & Hoffmann (2004); Gerevini & Long, (2005); Fox & Long, (2006); Helmert, (2008); PDDL 3.1, PDDL +) family of languages developed in the planning community, which in turn took inspiration from STRIPS (Fikes & Nilsson, 1971) and the situation calculus (McCarthy and Hayes 1950).

While we share the objective of representing environments simply and concretely, our primary objective is to enable a complete description of a range of transformation operators, which necessitates a different set of domain representation decisions. This results in a similar representation to PDDL languages that is not directly descended from any one such language. Ease of planning was not a primary design criterion, and we do not attempt to provide backwards compatibility with existing PDDL languages; instead, T-SAL follows the same guidelines but without historical baggage.

In describing T-SAL domains, we use symbols as identifiers; we refer to the set of symbols as *S*. Variables in a domain description come from the set *V* ⊂ *S* (the space of variable sets is denoted **V**). Another set of symbols identify particular objects; we say object *o* comes from the set *O* ⊂ *S*.

⌐ A domain type *ty* is in the space *Ty* ⊂ *S.* Five symbols are reserved for special basic types: REAL, INTEGER, BOOLEAN, AGENT, and OBJECT. A sixth standard type, POSITION, is also reserved, but is not a "basic" type. Nearly every domain requires objects and sometimes agents to be related spatially, and spatial information is useful to distinguish from non-spatial. Therefore, POSITION is reserved as a type to allow spatial properties to be automatically distinguished from non-spatial properties. The POSITION type must be specified to derive from REAL, INTEGER, or OBJECT, depending on the representation of a domain.

**Definition 1: Domain Types** ⌐

>    *Example*: Unique types in CartPole include CART, BLOCK, SPEED, and ANGLE.

⌐ A domain function uses the variable *f* and has space *F* and space of sets **F**, has a name (*function-name*: *F* → *Fn*), arguments (*function-arguments*: *F* → **V**) and types (*function-argument-type*: *F* × *V* → *Ty*), and a value type (*function-value-type*: *F* → *Ty*). Function names *fn* come from the set *Fn* ⊂ *S*. In tuple form, a function is given as ⟨*name*, *arguments*, *argument-types*, *value-type*⟩.

**Definition 2: Domain Functions and Function Names** ⌐

>    *Example*: In T-SAL, the relationship between the cart and pole are described using roll, pitch, and yaw. The function *f* below describes the roll of a pole relative to its cart in T-SAL-CR:
>
>    *f*: (CART-VELOCITY ?C – CART) – SPEED
>
>    For this function, the following equivalences hold: *function-name*(*f*) = CART-VELOCITY; *function-arguments*(*f*) = [?C]; *function-argument-type*(*f*, ?C) = CART; *function-value-type*(*f*) = SPEED.

⌐ Values include objects, numbers, and Boolean literals; the set of values is defined *Va* ≡ *O* ∪ 𝕀 ∪ ℝ ∪ {TRUE, FALSE}. Terms include values, variables, and function terms. Function terms apply a function to other terms, including function terms. Formally, the set of terms is defined *Te* ≡ *V* ∪ *Va* ∪ ⟨*Fn*, **Te**⟩. Here, **Te** is the space of sequences of terms from *Te*. Legal function terms have a term sequence the same length as the arity of the named function.

**Definition 3: Values and Terms** ⌐

>    *Examples*: In the T-SAL domain language representation of Cart-Pole, legal terms include 30, CART1, ?C, (POLE-ANGLE ?C), and (POLE-ANGLE (LEFTMOST-CART)).

| Concept | *Var, Space* | L(T-SAL) Definition | T-SAL-CR Example |
|---|---|---|---|
| Symbol | S | <enumerated class> | CART |
| Variable | v ∈ V | V ⊂ S | ?C |
| Term | Te | Te ≡ V ∪ Va ∪ ⟨Fn, **Te**⟩ | ?C; CART; 3; 1.5; TRUE; (CART-VELOCITY ?C) |
| Type | ty ∈ Ty | Ty ⊂ S | CART |
| Function | f ∈ F | F ≡ S × **V** × (V → S) × S | (CART-VELOCITY ?C — CART) — SPEED |
| Ground Fluent | gf ∈ GF | GF ≡ Fn × (V → Va) × Va<br>Tuple: ⟨fn, args, value⟩ | (= (CART-VELOCITY CART1) 0) |
| Calculation | calc ∈ Ca | Ca ≡ Te ∪ F × **Te** ∪ Op × Ca × Ca<br>∪ AggOp × V × Cn × Ca<br>∪ IF Cn × Ca × Ca<br>Tuples:<br>⟨fun, terms⟩ ⟨op, calc1, calc2⟩<br>⟨SUM, v, con, calc⟩ ⟨PRODUCT, v, con, calc⟩ ⟨IF, cnd, calc_T, calc_F⟩ | 0; (POLE-ANGLE ?C);<br>(+ (POLE-ANGLE ?C) .01);<br>(/ (SUM ?C (≠ (CART-VELOCITY ?C) 0)<br>    (POLE-ANGLE ?C))<br>  (SUM ?C (≠ (CART-VELOCITY ?C) 0) 1)) |
| Condition | cnd ∈ Cn | Cn ≡ Te ∪ PC ∪ BOp × Cn × Cn<br>∪ FORALL × **V** × Cn × Cn<br>∪ HAS-VALUE × V × Ca<br>Tuples:<br>⟨AND, cnd1, cnd2⟩ ⟨OR, cnd1, cnd2⟩<br>⟨FORALL, **v**, cons, req⟩ ⟨HAS-VALUE, v, calc⟩ | TRUE; (< (POLE-ANGLE ?C) 30);<br>(AND (< (POLE-ANGLE CART1) 30) (> (POLE-ANGLE CART1) -30));<br>(FORALL ?C (≠ (CART-VELOCITY ?C) 0)<br>  (< (POLE-ANGLE ?C) 30));<br>(HAS-VALUE ?A (POLE-ANGLE ?C)) |
| Effect | e ∈ E | E ≡ S × **V** × Alt × Te × [0..1] | (DECREASE (POLE-ANGLE ?C) 0.1) [0.5] |
| Continuous Change | c ∈ C | C ≡ S × **V** × Dir × Ca | (INCREASE (POLE-ANGLE ?C) (* DT<br>  (ANGULAR-MOTION ?C)) |
| Action Model | a ∈ A | A ≡ S × Te × **V** × (V → Ty) × **Cn** × **E**<br>Tuple:<br>⟨name, performer, args, argTypes, precs, effs⟩ | (:ACTION PUSH<br> :PERFORMER ?AG<br> :PARAMETERS (?C — CART)<br> :PRECONDITIONS ((CONTROLS ?AG ?C)<br>              (= (AGENT-FORCE ?AG))<br> :EFFECTS<br>  ((INCREASE (CART-VELOCITY ?C) ?FORCE)) |
| Event Model | ev ∈ Ev | Ev ≡ S × **V** × (V → Ty) × [0..1] × ℝ^{≥0} × **Cn** × **E**<br>Tuple:<br>⟨name, quals, qualTypes, prob, freq, triggers, effs⟩ | (:EVENT FINISHES<br> :QUALITIES (?C — CART)<br> :TRIGGERS ((UPRIGHT ?C)<br>        (< (CART-VELOCITY ?C) 0.1)<br>        (> (CART-VELOCITY ?C) -0.1)<br>        (CONTROLS ?AG ?C))<br> :EFFECTS (WINS ?AG)) |
| Process Model | p ∈ P | P ≡ S × **V** × (V → Ty) × **Cn** × **C**<br>Tuple:<br>⟨name, quals, qualTypes, conds, changes⟩ | (:PROCESS POLE-ANGLE-CHANGES<br> :QUALITIES (?C — CART)<br> :CONDITIONS (FORALL ?AG TRUE<br>             (NOT (WINS ?AG)))<br> :EFFECTS<br>  (AND (INCREASE (POLE-ANGLE ?C)<br>      (* DT (ANGULAR-MOTION ?C))))) |

Table 2 – Summary of key T-SAL domain concepts with definitions and examples

⌜ Conditions (variable *cnd*, space *Cn,* space of sequences ***Cn***) specify complex requirements over states using comparisons and Boolean operators. Like calculations, conditions have the form of a term, a comparison of two values, or a combined calculation in which two operands are combined by a Boolean operation: *Cn ≡ Te ∪ BOp × Cn × Cn ∪ FORALL × V × Cn × Cn ∪ Ineq × Ca × Ca ∪ NOT × Cn*. Supported Boolean operations are *BOp ≡* {AND, OR}. Supported comparison operators are the inequalities *Ineq ≡* {=, ≠, <, >, ≤, ≥}.

**Definition 4: Conditions** ⌟

*Examples*: In the T-SAL knowledge representation of Cart-Pole, legal conditions include:

- (< (POLE-ANGLE ?C) 30)
    - True iff the pole on the cart bound to ?C has roll less than 30
- (AND (< (POLE-ANGLE CART1) 30) (> (POLE-ANGLE CART1) -30))
    - True iff the pole on the object CART1 has angle between -30 and 30
- (AND (<= (POLE-ANGLE CART1) (POLE-ANGLE CART2)))
    - True iff the pole on the object CART1 has a smaller angle than that on CART2.
- (FORALL ?C (!= (CART-VELOCITY ?C) 0) (< (POLE-ANGLE ?C) 30))
    - True iff every pole with non-zero velocity also has pole angle less than 30

⌜ Calculations (variable *calc*, space *Ca,* space of sets ***Ca***) are functional expressions composed of terms and operators. Calculation operators include basic arithmetic operations, distribution draws, aggregations over sets, and the trinary if operator. Where terms have the semantics of a lookup in a database incorporating state information, calculations describe executed procedures that return values. A calculation must be a term, a combined calculation in which two operands are combined by an operation, an aggregation over a set of variables, or a trinary if that returns the result of one of two sub-calculations: *Ca ≡ Te ∪ Op × Ca × Ca ∪ AggOp × V × Cn × Ca ∪ IF Cn × Ca × Ca*. Supported operations are *Op ∈* {+, - *, /, :UNIFORM, :GAUSSIAN}. These operations have the usual definitions, and all require that their operands are numeric (i.e., derived from the type REAL or INTEGER). The +, -, *, and / operations yield a value of type REAL if either operand is REAL, or INTEGER otherwise. The :UNIFORM operation draws a value from the uniform distribution, with a resulting value in the range defined by its operands (the first operand providing the lower bound, and the second the upper bound, inclusive). The operands and the resulting value must have the INTEGER type. The :GAUSSIAN operation draws from the gaussian distribution; its operands are interpreted as a mean and standard distribution, and the type of the calculation will be REAL. Aggregation operations are *AggOp ∈* (SUM, PRODUCT). Aggregations have a single variable, constrained to meet a condition; the sub-calculation is calculated for every permissible value of the variable and results are aggregated. Conditional calculations give a condition that determines which one of a pair of calculations to evaluate; if the condition evaluates to TRUE, the first calculation is evaluated, and if FALSE the second.

**Definition 5: Calculations** ⌟

*Example*: In the T-SAL knowledge representation of Cart-Pole, legal calculations include:

- (POLE-ANGLE ?C)
- (+ (POLE-ANGLE ?C) .01)
- (/ (SUM ?C (≠ (CART-VELOCITY ?C) 0) (POLE-ANGLE ?C)) (SUM ?C (≠ (CART-VELOCITY ?C) 0)  1))
    - Average angle of poles with nonzero velocity

⌐ Axioms define the truth of fluents in a state in terms of a condition. A fluent defined by axioms must have value type BOOLEAN and takes on the value TRUE when the associated condition is satisfied, or FALSE otherwise. No function may therefore both have an axiom definition and appear in action or event effects. Axioms use the variable *ax* (*Ax* for the space, ***Ax*** for the space of sets), and describes a function name (*axiom-name*: *Ax* → *S*), argument types (*axiom-argument*: *Ax* × *V* → *S*), and an antecedent condition (*axiom-antecedent*: *Ax* → *Cn*).

**Definition 6: Axioms** ⌐

> *Example*: In Cart-Pole, an axiom *ax* could say that a "live" cart is one whose pole angles are less than 30:
>
> (:- (UPRIGHT ?C - CART)
>   (AND (< (POLE-ROLL ?C) 0.1) (> (POLE-ANGLE ?C) -0.1)))
>
> Here, *axiom-name*(*ax*) = UPRIGHT; *axiom-argument*(*ax*, ?C) = CART; and *axiom-antecedent*(*ax*) = ⟨AND, ⟨<, POLE-ANGLE, [?C], 0.1⟩ ⟨>, POLE-ANGLE, [?C], -0.1⟩⟩

⌐ An effect, which uses the variable *e* and has space *E* (***E*** for the space of sets), describes an instantaneous alteration to the world caused by an action or effect transition. Mostly, these change the values of fluents. However, a second type of effect creates a new object; this is necessary in general to allow us to describe open worlds where objects can proliferate without a bound. An effect has a symbol that names the function it affects (*effect-name*: *E* → *Fn*), assignments to that function's arguments (*effect-argument*: *E* × *V* → *Te*), a modification type (*effect-modification*: *E* → *Alt*), post-effect value (*effect-value*: *E* → *Te*), and probability (*effect-probability*: *E* → [0..1]). The set of possible alterations is given by: *Alt* ≡ {SET, INCREASE, DECREASE, CREATE}. In the special case of a CREATE alteration, the effect's name is the type symbol of the object created rather than that of a function, and the effect's value describes a name prefix to use in referring to the new object.

**Definition 7: Effects** ⌐

> *Example*: In Cart-Pole, an example legal effect *e* is (DECREASE (POLE-ANGLE ?C) 0.1) [0.5], which states that the value of the fluent POLE-ANGLE for the object bound to ?C decreases by 0.1 during a transition. Here, *effect-name*(*e*) = POLE-ANGLE; *effect-argument*(*e*, ?C) = ?C; *effect-modification*(*e*) = DECREASE; *effect-value*(*e*) = 0.1; and *effect-probability*(*e*) = 0.5.
>
> A CREATE effect of the form (CREATE BLOCK ?B "new-block") causes a new block to exist in the Cart-Pole environment. Here, *effect-name*(*e*) = BLOCK; *effect-argument*(*e*, BLOCK) = ?B; *effect-modification*(*e*) = CREATE; and *effect-value*(*e*) = "new-block". With no probability explicitly described, we give *effect-probability*(*e*) = 1.

⌐ A continuous change, which uses the variable *c* and has space *C* (***C*** for the space of sets), describes continuous time changes to fluents. A change describes the fluent it changes with a name symbol (*change-name*: *C* → *S*), argument assignments to that fluent (*change-argument*: *C* × *V* → *Te*), direction of change (*change-direction*: *C* → *Dir*) and a derivative calculation (*change-derivative*: *C* → *Ca*). Here, the set of possible directions is given by: *Dir* ≡ (increase, decrease). A special DT symbol is reserved to represent the time differential in continuous change derivatives only. Each continuous change must use this symbol once.

**Definition 8: Continuous Changes** ⌐

*Example*: In Cart-Pole, an example legal change *c* is (INCREASE (POLE-ANGLE ?C) (* DT (ANGULAR-MOTION ?C))), which states that the value of the fluent POLE-ANGLE for the cart bound to ?C increases linearly with time according to the angular motion function (not described further here). Here, *change-name*(*c*) = POLE-ANGLE; *change-argument*(*c*, ?C) = ?C; *change-direction*(*c*) = INCREASE; and *change-derivative*(*c*) = (* DT (ANGULAR-MOTION ?C)).

⌜ Action models, event models, and process models describe when and how the world changes. Actions represent change caused by a performer; a performer can choose whether to take an action and with what parameters. Some actions may act differently for different performers or not be available to certain performers, so a performer variable gives a reference to the performer for preconditions and effects. Formally, an action model $a \in A$ (space of sets **A**) has a name (*action-name*: $A \to S$), performer (*action-performer*: $A \to Te$), parameters (*action-parameters*: $A \to V$), parameter types (*action-parameter-type*: $A \times V \to Ty$), preconditions (*action-preconditions*: $A \to Cn$), and effects (*action-effects*: $A \to E$). Tuple notation for actions is ⟨*name, performer, args, argTypes, precs, effs*⟩; a tuple in this form can be interpreted as an action that can respond to each above function.

**Definition 9: Action Models** ⌟

*Example*: An action model *a* in cart pole pushes the cart forward, instantaneously increasing its velocity in the x direction:

(:ACTION PUSH
 :PERFORMER ?AG
 :PARAMETERS (?C – CART)
 :PRECONDITIONS ((CONTROLS ?AG ?C) (= (AGENT-FORCE ?AG) ?FORCE)
 :EFFECTS ((INCREASE (CART-VELOCITY ?C) ?FORCE))

Here, *action-name*(*a*) = PUSH; *action-performer*(*a*) = ?AG; *action-parameters*(*a*) =[?C]; *action-parameter-type*(*a*, ?C) = CART; *action-preconditions*(*a*) = [(= (AGENT-FORCE ?AG) ?FORCE) (CONTROLS ?AG ?C)]; and *action-effects*(*a*) = [(INCREASE (CART-VELOCITY ?C) 1)].

⌜ Event models represent instantaneous change not caused by a modelled agent; they have no parameters, because no entity (performer or otherwise) chooses an event to happen. However, qualities refer to a set of variables that may be instantiated within the triggers of the event and aid in understanding. All events are triggered in one of three ways: occur 1) by default, they must occur immediately when their triggers are met; 2) if an event model has a probability between 0 and 1, modelled events occur probabilistically when their triggers are met; 3) if an event model has a non-zero frequency, modelled events occur at random times, with interarrival times described by a poisson distribution with the modelled frequencyNo event model can have both a probability less than 1 and a non-zero frequency. An event $ev \in Ev$ (space of sets **Ev**) has a name (*event-name*: $Ev \to S$), qualities (*event-qualities*: $Ev \to V$), quality types (*event-quality-type*: $Ev \times V \to Ty$), probability (*event-probability*: $Ev \to [0..1]$), frequency (*event-frequency*: $Ev \to \mathbb{R}^{\geq 0}$), triggers (*event-triggers*: $Ev \to Cn$), and effects (*event-effects*: $Ev \to E$). Tuple notation for events is ⟨*name, quals, qualTypes, prob, freq, triggers, effs*⟩; a tuple in this form can be interpreted as an event that can respond to each above function.

**Definition 10: Event Models** ⌟

*Example*: An event *ev* in cart pole causes a win for a player whose cart reaches an angle and velocity within a certain range:

(:EVENT FINISHES
 :QUALITIES (?C – CART)
 :TRIGGERS ((UPRIGHT ?C) (< (CART-VELOCITY ?C) 0.1) (> (CART-VELOCITY ?C) -0.1) (CONTROLS ?AG ?C))
 :EFFECTS (WINS ?AG))

Here, *event-name*(*ev*) = FINISHES; *event-probability*(*ev*) = 1; *event-frequency*(*ev*) = 0; *event-qualities*(*ev*) =[?C]; *event-quality-type*(*ev*, ?C) = CART; *event-triggers*(*ev*) = [(= (UPRIGHT ?C) TRUE), (< (CART-VELOCITY ?C) 0.1), (> (CART-VELOCITY ?C) -0.1), (= (CONTROLS ?AG ?C) TRUE)]; and *event-effects*(*ev*) = [(SET (WINS ?AG) TRUE)].

⌐ Process models represent continuous change over time. Like events, they have qualities and are not controlled by any agent. Unlike events (or actions), they represent change over time, and describe continuous changes that occur over the duration of the process rather than effects. A process will continue for as long as its conditions are met. A process $p \in P$ (space of sets **P**) has a name (*process-name*: $P \to S$), qualities (*process-qualities*: $P \to V$), quality types (*process-quality-type*: $P \times V \to Ty$), conditions (*process-conditions*: $P \to Cn$), and continuous changes (*process-changes*: $P \to C$). Tuple notation for processes is ⟨*name, quals, qualTypes, conds, changes*⟩; a tuple in this form can be interpreted as a process model that can respond to each above function.

**Definition 11: Process Models** ⌐

*Example*: A process *p* in cart pole causes the pole to change position:

(:PROCESS POLE-ANGLE-CHANGES
 :QUALITIES (?C – CART)
 :CONDITIONS (FORALL ?AG TRUE (NOT (WINS ?AG)))
 :EFFECTS (AND (INCREASE (POLE-ANGLE ?C) (* DT (ANGULAR-MOTION ?C)))))

Here, *process-name*(*p*) = POLE-ANGLE-CHANGES; *process-qualities*(*p*) =[?C]; *process-quality-type*(*p*, ?C) = CART; *process-conditions*(*p*) = [(FORALL ?AG TRUE (= (WINS ?AG) FALSE))]; and *process-changes*(*p*) = [(INCREASE (POLE-ANGLE ?C) (* DT (ANGULAR-MOTION ?C)))].

⌐ A T-SAL domain, which we use the variable *d* for (*D* is the space of possible domains), includes types (*domain-types*: $D \to S$), supertypes (*domain-supertypes*: $D \to S \times S$), named constants with given types (*domain-constants*: $D \to S \times S$), functions (*domain-functions*: $D \to F$), axioms (*domain-axioms*: $D \to Ax$), actions (*domain-actions*: $D \to A$), events (*domain-events*: $D \to E$), and processes (*domain-processes*: $D \to P$). Tuple notation for domains is ⟨*types, supertypes, constantTypes, functions, axioms, actions, events, processes*⟩.

**Definition 12: T-SAL Domains** ⌐

Legality of a T-SAL domain, based mainly on type agreement, is described in Appendix A.

## 3.2 T-SAL States, Goals, and Scenario Generators

In addition to dynamics, T-SAL describes environments in terms of distributions of starting states and tasks. In this section, we define T-SAL scenario generators, which define both of these. To describe a scenario generator, we need the concepts of state, and various simpler generator functions: object generators, value generators, and fluent generators.

⌐ A T-SAL state *st* describes what is true in the environment at a particular instant in time. We describe a state with a tuple ⟨*objects*, *defaults*, *assignments*⟩. Each T-SAL state is associated with a T-SAL domain *d* = ⟨*types, supertypes, constantTypes, functions, axioms, actions, events, processes*⟩, and defines the current set of objects belonging to every type in *types* and a value for every possible fluent grounding using *functions* with those objects. State objects (*state-objects*: $St \times S \rightarrow S$) includes all objects of every type in the domain. The defaults (*state-defaults*: $St \times F \rightarrow Te$) give a default value for every function, which is assigned by this state to any fluent grounding that is not given a specific value in *assignments*. The state's assignments (*state-assignments*: $St \rightarrow \mathbf{GF}$) give specific values for a list of ground fluents. Ground fluents use the variable *gf*, space *GF* and space of sequences **GF**. Each ground fluent names the function the fluent is from (*ground-fluent-function-name*: $GF \rightarrow Fn$), argument values (*ground-fluent-argument*: $GF \times V \rightarrow Va$) and a fluent value (*ground-fluent-value*: $GF \rightarrow Va$). Tuple form for ground fluents is ⟨*function-name*, *argument-values*, *fluent-value*⟩.

**Definition 13: T-SAL State** ⌐

> *Example*: In a simple Cart-Pole state *st* there is one cart, CART1, and three blocks, BLOCK1, BLOCK2, and BLOCK3. A single agent, AGENT1 exists. Other types (speed, position, angle) are derived from REAL and are non-enumerated. Fluents in the domain include *f1* = (CART-VELOCITY ?C – CART) – SPEED, *f2* = (CART-POSITION ?C – CART) – POSITION, *f3* = (POLE-ANGLE ?C – CART) – ANGLE, *f4* = (BLOCK-POSITION ?B – BLOCK) – POSITION, *f5* = (BLOCK-VELOCITY ?B – BLOCK) – SPEED, *f6* = (AGENT-FORCE ?AG – AGENT) – REAL, *f7* = (CONTROLS ?AG – AGENT ?C - CART) – BOOLEAN, and *f8* = (WINS ?AG – AGENT) – BOOLEAN. Relevant values are: *state-objects*(*st*, CART) = [CART1]; *state-objects*(*st*, BLOCK) = [BLOCK1, BLOCK2, BLOCK3]; *state-defaults*(*st*, *f1*) = 0; *state-defaults*(*st*, *f3*) = 0.1; *state-defaults*(*st*, *f5*) = 0; *state-defaults*(*st*, *f6*) = 1; *state-defaults*(*st*, *f7*) = FALSE; *state-defaults*(*st*, *f8*) = FALSE; *state-assignments*(*st*) = [CART-POSITION(CART1) = 0, BLOCK-POSITION(BLOCK1) = 5, BLOCK-POSITION(BLOCK2) = -5, BLOCK-POSITION(BLOCK3) = 10, CONTROLS(AGENT1, CART1) = TRUE].

> Note that every legal assignment to every fluent can be valued using either a default or an explicit assignment in *st*. In this case, functions *f2* and *f4*, which have no defaults, must be fully enumerated, as shown.

⌐ An object generator describes a distribution of objects in the environment. Whereas a T-SAL domain refers to certain "constant" objects present in every scenario, object generators describe the objects that are present in an environment as part of a particular scenario. An object generator *og* has a name (*object-generator-name*: $OG \rightarrow S$), type (*object-generator-type*: $OG \rightarrow S$), and a draw function that yields a set of distinct objects when called (*object-generator-draw*: $OG \rightarrow (\rightarrow \mathbf{S})$). As randomness is frequently a desirable property of such generators, the draw function may returns sets with different contents and sizes on subsequent invocations. A set of basic draw functions is described in Section 3.4.3.

**Definition 14: Object Generator** ⌐

*Example*: A simple object generator *og* makes 5 block objects. We write this like so:

> OBJECTGENERATOR(blockGroup, BLOCK, OBJECTLIST(5, "block"))

The standard draw function OBJECTLIST is defined to return *N* (the first argument) object names, each starting with the prefix given by the second argument. The object generator *og* has the following characteristics: *object-generator-name*(*og*) = cartGroup; *object-generator-type*(*og*) = CART; *object-generator-draw*(*og*) = OBJECTLIST(5, "cart").

⌐ A value generator returns values from a random distribution; these values may be generated by from sets defined by object generators, domain constants, or from the reals, integers, or booleans. A value generator *vg* has a name (*value-generator-name*: VG → S) and provides a draw function that yields a sequence of values when called (*value-generator-draw*: VG → (→ ***Va***)). The draw function may produce random results or static results; a set of basic value generator draw functions is given in Section 3.4.4.

**Definition 15: Value Generator** ⌐

*Example*: A simple value generator *vg* for positions gives a single random real value between -20 and 20. We write this like so:

> VALUEGENERATOR(RANDOMPOSITION, UNIFORMREALDISTRIBUTION(-20, 20))

This uses a standard draw function that pulls from a uniform distribution. The value generator *vg* has the following characteristics: *value-generator-name*(*vg*) = RANDOMPOSITION; *value-generator-draw*(*og*) = UNIFORMREALDISTRIBUTION(-20, 20) .

⌐ A fluent generator describes a distribution of fluent values for a particular domain function. A fluent generator *fg* has a function name (*fluent-generator-function-name*: FG → Fn) that it generates fluents for, and provides a draw function (*fluent-generator-draw*: FG → (→ ***GF***)) that yields a list of ground fluents describing individual fluent groundings; these do not necessarily map all possible assignments to the named function. Basic fluent generators are described in Section 3.4.5.

**Definition 16: Fluent Generator** ⌐

*Example*: A simple fluent generator *fg* for assigning block positions gives a random real value between -20 and 20 to each block. We write this like so:

> FLUENTGENERATOR(BLOCK-POSITION, ALLPERMUTATIONS([BLOCKGROUP], RANDOMPOSITION)

The standard draw function, ALLPERMUTATIONS, described here creates a ground fluent for every permutation of unique objects generated by the object generators given as its first argument, and assigns each a value generated by the value generator given as its second argument. The fluent generator has the following characteristics: *fluent-generator-function-name*(*fg*) = BLOCK-POSITION; *fluent-generator-draw*(*fg*) = ALLPERMUTATIONS([BLOCKGROUP], RANDOMPOSITION).

⌐ A T-SAL scenario generator *sg* is a tuple ⟨*fluents*, *defaults*, *object-generators*, *value-generators*, *fluent-generators, performance*⟩, that describes a distribution over environment starting states, as well as a performance calculation that describes the success of each agent. The first item, *fluents*, (*scenario-* *generator-object-generators*: *SG* → **GF**), gives a set of ground fluents present in every scenario generated; these are not randomized. The second item, *defaults*, (*scenario-generator-defaults*: *SG* × *Fn* → *Va*) maps each function name to a default value that can be assumed in the absence of contradictory assignments (from the fluents list and fluent generators). Each fluent in the corresponding domain should be covered by this mapping. The third item, *object-generators*, (*scenario-generator-object-generators*: *SG* → **OG**) sets up objects present in a starting state. The fourth item, *value-generators*, (*scenario-generator-value-generators*: *SG* → **VG**) gives a set of value generators defined for use by the fluent generators. The fifth scenario generator item, *fluent-generators,* (*scenario-generator-fluent-generators*: *SG* → **FG**) gives a set of fluent generators that describe distributions over fluents in the starting state. Finally, the last item, *performance*, gives a performance calculation as a calculation (*scenario-generator-performance-calculation*: *SG* → *Ca*). This calculation must have a single free variable, ?AG, which identifies the agent for which performance is to be computed. For simplicity, we assume that this calculation is always intended to be maximized and can be evaluated in any state.

**Definition 17: T-SAL Scenario Generator** ⌋

> *Example*: A scenario generator *sg* defines cart-pole performance as the negative sum of the squared angles of all poles controlled by that agent.
>
> *scenario-generator-performance-calculation(sg) =*
> (SUM ?C (CONTROLS ?AG ?C) (- (* (POLE-ANGLE ?C) (POLE-ANGLE ?C))))

To generate a scenario, a new state object is created with defaults from the state generator, objects created by drawing from each object-generator, and assignments created by drawing from each fluent generator. Value generators are used indirectly during fluent generators draws. Finally, fluents in the scenario generator fluents list are added to the state, overwriting any generated and/or default fluents with the same fluent name and arguments.

An environment consists of a domain *d* and a scenario generator *sg*. To be legal, *sg* must have a close relationship to the domain *d*:

- *fluents* must have only ground fluents with appropriate arity and values of appropriate types
- *defaults* must give an appropriately typed value for every function in *d*
- *object-generators* must use object types in *d*
- *fluent-generators* must always generate legal ground fluents for the described function, which must also be in domain *d*
- *performance* must be a real-valued calculation with a single free variable, ?AG, referring to the target agent, it must only reference functions in the domain *d*, and all symbols must reference constants from the domain *d*.

Since the returns from generator draw functions are defined to vary over time, constraints on the draw function outputs are expressed below using the "G" temporal operator, which means "always".

## 3.3 T-SAL Transformations and Transformation Sequences

⌜ We use the variable *t* to describe an environment transformation, given as a tuple with a head and named arguments. The space of possible environment transformations is denoted *T*. We will also refer to a transformation sequence ***t*** = [***t***$_0$, …, ***t***$_n$], with the transformation sequence space denoted as ***T***. A transformation has a type (*transformation-type*: $T \rightarrow S$) and arguments (*transformation-argument*: $T \times S \rightarrow U$). Here, *U* denotes the universal set; the meaning of particular transformation arguments is dependent on the type. Individual transformations are defined in Section 3.5.2 (T-Transformations) and Appendix B (R-Transformations).

**Definition 18: Transformations and Transformation Sequences** ⌟

*Example*: A transformation *t* to a cart-pole environment that subtracts push force from a budget is given as:

ADDACTIONEFFECT(ACTIONNAME: PUSH, EFFECT: (DECREASE (PUSH-BUDGET ?AG) ?FORCE))

The transformation *t* has *transformation-type*(*t*) = ADDACTIONEFFECT. The first argument is referred to by *transformation-argument*(*t*, ACTIONNAME) = PUSH.

⌜ Applying a transformation to an environment yields a new environment: *apply*: $T \times D \times SG \rightarrow D \times SG$. We also define a function that applies a sequence of transformations (*applySequence*: ***T*** $\times D \times SG \rightarrow D \times SG$) as follows.

*applySequence*(∅, *d, sg*) ≡ ⟨*d, sg*⟩
*applySequence*(***t*** = [***t***$_0$, …, ***t***$_n$], *d, sg*) ≡ *applySequence*([***t***$_1$, …, ***t***$_n$], *apply*(*t*$_0$, *d, sg*)) .

**Definition 19: Transformation and Transformation Sequence Application** ⌟

## 3.4 T-SAL R-Transformations

R-transformations are modifications to the environment dynamics. We formally define 30 transformations on T-SAL environments in Appendix A with a summary of these given here in Table 3. State space transformations are those that change the set of states the environment may contain. Performer control transformations change a performer's ability to interact with the environment. Instanteous environment change transformations change all non-voluntary instanteous transitions. Durative natural change transformations change processes which operate over continuously over time.

| Summary of all R-transformations | | | |
|---|---|---|---|
| **State Space Transformations** | **Performer Control Transformations** | **Instanteous Environment Change Transformations** | **Durative Natural Change Transformations** |
| ADDTYPE | ADDACTION | ADDEVENT | ADDPROCESS |
| ADDTYPEPARENT | ADDPRECONDITION | CHANGEFREQUENCY | ADDPROCESSCONDITION |
| ADDCONSTANT | ADDACTIONEFFECT | CHANGEPROBABILITY | ADDPROCESSCHANGE |
| ADDFUNCTION | REMOVEACTION | ADDTRIGGER | REMOVEPROCESS |
| ADDAXIOM | REMOVEPRECONDITION | ADDEVENTEFFECT | REMOVEPROCESSCONDITIONS |
| REMOVETYPE | REMOVEACTIONEFFECT | REMOVEEVENT | REMOVEPROCESSCHANGE |
| REMOVETYPEPARENT | | REMOVETRIGGER | |
| REMOVECONSTANT | | REMOVEEVENTEFFECT | |
| REMOVEFUNCTION | | | |
| REMOVEAXIOM | | | |

Table 3: The *apply* function is defined with respect to all these R-Transformations within Appendix A.

## 3.5 T-SAL T-Transformations

T-Transformations are modifications to the scenario generator. We define a set of T-Transformations which cover the space of possible generator modifications below; first, we give examples for a paradigmatic domain, then we define the particular transformations possible, as well as the sub-generators that can be used.

To easily transform scenario generators, it will be useful to supersede existing value or object generators. Therefore, when we add a new generator, it should be assumed to replace any existing generator with the same name, and all references in other generators to the old generator become references to the new generator. As such, we need only five types of T-Transformation: ADDFLUENTVALUE, ADDDEFAULTVALUE, ADDOBJECTGENERATOR, ADDVALUEGENERATOR, ADDFLUENTGENERATOR, and ADDPERFORMANCECALCULATION. As they are particularly novel, we give additional examples below of T-Transformations for value, object, and fluent generators useful in various domains; we follow this with a formal definition of these five transformations, and a description of the value, object, and fluent generators used.

### 3.5.1 T-Transformation examples

In Mudworld (Molineaux and Aha 2014), a set of rovers attempt to travel to a set of destinations on a 6x6 grid. Any location may be muddy, with a 30% chance. Mud slows the rover down to half speed, with some variability across scenarios. All destinations are within four cell moves. The following T-Transformations construct a scenario generator for this environment.

ADDVALUEGENERATOR(XLoc, UNIFORMINTEGERDISTRIBUTION(1, 6))
ADDVALUEGENERATOR(Yloc, UNIFORMINTEGERDISTRIBUTION(1, 6))
ADDVALUEGENERATOR(Xset, INTEGERSEQUENCE(1, 6))
ADDVALUEGENERATOR(Yset, INTEGERSEQUENCE (1, 6))
ADDOBJECTGENERATOR(robotGroup, Robot, OBJECTLIST(6, "rover"))
ADDVALUEGENERATOR(RobotG, DRAWFROMOBJECTSET(robotGroup))
ADDVALUEGENERATOR(RandRobot, DRAWALLFROMOBJECTSET(robotGroup))
ADDVALUEGENERATOR(TRUE, CONSTANTFUNCTION(TRUE))

ADDVALUEGENERATOR(IsMuddy, BERNOULLIDISTRIBUTION(0.3))
ADDFLUENTGENERATOR("muddy", ALLPERMUTATIONS([Xset, Yset], IsMuddy))
ADDVALUEGENERATOR(SpeedInMud, GAUSSIANDISTRIBUTION(0.5, 0.2))
ADDFLUENTGENERATOR("speed-in-mud", ALLPERMUTATIONS([], SpeedInMud))
ADDVALUEGENERATOR(ClosePos, FILTER(
                                  DRAWTUPLE([Xloc, Yloc, Xloc, Yloc], ["x1", "y1", "x2", "y2"])),
                                  $\lambda(x_1, y_1, x_2, y_2) \rightarrow \{|x_1 - x_2| + |y_1 - y_2| < 4\}$)))
ADDVALUEGENERATOR(CloseSet, NEWSET(NDRAWS(ClosePos, 6)))
ADDFLUENTGENERATOR("robot-x-loc", NFLUENTDRAWS([Robot], CloseSet.x1, 6))
ADDFLUENTGENERATOR("robot-y-loc", NFLUENTDRAWS([Robot], CloseSet.y1, 6))
ADDFLUENTGENERATOR("robot-dest", NFLUENTDRAWS([Robot, CloseSet.x2, CloseSet.y2], TRUE, 6))

Here, we consider a scenario generator that creates a multi-dimensional random value for a camera's starting location:

ADDVALUEGENERATOR(Position, NDIMENSIONALGAUSSIANDISTRIBUTION([0.5, 0.7, 0.2], [0.1, 0.2, 0.05]))
ADDFLUENTGENERATOR("camera-starting-position",
                    NFLUENTDRAWS([Position.x1, Position.x2, Position.x3], True, 1))

### 3.5.2 T-Transformation Definitions

We define the apply function for five T-Transformations:

*apply*(ADDFLUENTVALUE(FUNCTIONNAME: *name*, FLUENTARGS: args, VALUE: *value*), *d*,
      ⟨*fluents, defaults, object-generators, value-generators, fluent-generators, performance*⟩) ≡
  ⟨{⟨*fname, fargs, fvalue*⟩ | ⟨*fname, fargs, fvalue*⟩ ∈ *fluents* ∧ (*fname ≠ name* ∨ *fargs ≠ args*)}
     ∪ {⟨*name, args, value*⟩},
    *defaults, object-generators, value-generators, fluent-generators*⟩

*apply*(ADDDEFAULTVALUE(FUNCTIONNAME: *name*, VALUE: *value*), *d*,
      ⟨*fluents, defaults, object-generators, value-generators, fluent-generators, performance*⟩) ≡
  ⟨*fluents*, {⟨*fname, fvalue*⟩ | ⟨*fname, fvalue*⟩ ∈ *defaults* ∧ *fname ≠ name*} ∪ {⟨*name, value*⟩},
    *object-generators, value-generators, fluent-generators*⟩

*apply*(ADDOBJECTGENERATOR (NAME: *name,* TYPE: *type*, DRAWFUNCTION: *drawFn*), *d*,
      ⟨*fluents, defaults, object-generators, value-generators, fluent-generators, performance*⟩) ≡
  ⟨*fluents, defaults*,
    {*og* ∈ *object-generators* | *object-generator-name*(*og*) ≠ *object-generator-name*(*newGenerator*)}
     ∪ {⟨*name, type, drawFn*⟩},
    *value-generators, fluent-generators*⟩

*apply*(ADDVALUEGENERATOR (NAME: *name,* DRAWFUNCTION: *drawFn*), *d*,
      ⟨*fluents, defaults, object-generators, value-generators, fluent-generators, performance*⟩) ≡
  ⟨*fluents, defaults, object-generators*,
    {*vg* ∈ *value-generators* | *value-generator-name*(*vg*) ≠ *value-generator-name*(*newGenerator*)}
     ∪ {⟨*name, drawFn*⟩}, *fluent-generators*⟩

*apply*(ADDFLUENTGENERATOR (NAME: *name*, DRAWFUNCTION: *drawFn*), *d*,
⟨*fluents*, *defaults*, *object-generators*, *value-generators*, *fluent-generators*, *performance*⟩) ≡
⟨*fluents*, *defaults*, *object-generators*, *value-generators*,
{*fg* ∈ *fluent-generators* | *fluent-generator-name*(*fg*)≠*fluent-generator-name*(*newGenerator*)}
∪ {⟨*name, drawFn*⟩}⟩

*apply*(REPLACEPERFORMANCECALCULATION(PERFORMANCE: *calculation*), *d*,
⟨*fluents*, *defaults*, *object-generators*, *value-generators*, *fluent-generators*, *performance*⟩) ≡
⟨*fluents*, *defaults*, *object-generators*, *value-generators*, *fluent-generators*, *calculation*⟩

### 3.5.3 Object Generator Draw Functions

The following simple example of an object generator draw function grounds our examples. Others, particularly with variable output, are possible.

OBJECTLIST(*count*, *id*) = () → [*gensym*(*id*, *i*)]$_{i=1..count}$

> Generates *count* objects of the requested *type*, prepended with the string in *id*, as created by the Lisp "gensym" function.

### 3.5.4 Value Generators

The following examples of value generator draw functions ground our examples. Others are possible.

UNIFORMDISTRIBUTION(*min, max*)

> Generator always returns a sequence containing a single real number drawn from a uniform distribution between min and max.

UNIFORMINTEGERDISTRIBUTION(*min, max*)

> Generator always returns a sequence containing a single uniform integer draw between min and max (inclusive).

GAUSSIANDISTRIBUTION(*mean, stdev*)

> Generator always returns a sequence containing a single draw from a one-dimensional Gaussian distribution with the given parameters.

NDIMENSIONALGAUSSIANDISTRIBUTION(*means[], stdevs[]*)

> Generator always returns a sequence containing a single draw from an n-dimensional Gaussian distribution with the given parameters, as a tuple with values named [x1, x2, …].

BERNOULLIDISTRIBUTION(*p*)

> Generator returns TRUE with probability *p*, and otherwise returns FALSE.

INTEGERSEQUENCE(*min, max*)

> Generator always returns a sequence containing all integers between min and max, in order.

DRAWFROMOBJECTSET(*objectGeneratorName*)

> Generator always returns a sequence containing a single draw from an object generator on the same scenario generator, referenced by name.

DRAWALLFROMOBJECTSET(*objectGeneratorName*)

> Generator always returns a sequence containing all objects that were produced from an object generator on the same scenario generator, referenced by name, in a random order.

CONSTANTFUNCTION(*value*)

> Generator always returns a sequence containing the single value passed in, i.e., the sequence [*value*].

NDRAWS(*valueGenerator, n*)

> Generator returns a sequence of *n* values drawn from *valueGenerator*.

DRAWTUPLE(*valueGenerators*[], *names*[])

> Generator returns a sequence with a single tuple with values drawn from *valueGenerators* and addressable by names given in *names.*

FILTER(*valueGenerator*, *filterFn* (Any) → Boolean)

> Generator returns a sequence with a single value; this is the first value returned by repeatedly drawing from *valueGenerator* until the *filterFn* reports TRUE for one.

NEWSET(*valueGenerator*)

> Generator always returns the same sequence, which is first populated by drawing from valueGenerator.

### 3.5.5 Fluent Generators

The following examples of fluent generator draw functions ground our examples. Others are possible.

ALLPERMUTATIONS(*argumentGenerators*[], *fluentValueGenerator*)

> Generator returns ground fluents with arguments populated by value generators corresponding to the names in *argumentGenerator*, and values populated by value generators corresponding to *fluentValueGenerator* (can be referenced by name or provide draw function directly).
>
> All permutations of a set of draws from the argument generators are used to generate a list of ground fluents. Values from *fluentValueGenerator* will be used in order, and a new set drawn when none are left, until every permutation is assigned a fluent value.

NFLUENTDRAWS(*argumentGenerators*[], *fluentValueGenerator, n*)

> Generator returns *n* ground fluents. The arguments of each are populated by value generators corresponding to the names in *argumentGenerators*, and the value of each is populated by the value generator corresponding to *fluentValueGenerator*. All values from a single draw of each argument and fluent value generator are used in order; if fewer than *n* values are returned by a generator, it will be called again until *n* values have been generated.

COMBINEFUNCTIONS(*fluentGenerators*[])

Results of each fluent generator are concatenated.

# 4  Application to Novelty Hierarchy

This section gives a novelty type definition for each level of the novelty hierarchy introduced by the DARPA SAIL-ON program (see Table 2). These definitions were an early attempt to categorize transformations possible in open-world environments, starting with the properties of individual world elements (objects, agents, and their actions and goals), advancing to multi-element properties (relations, interactions, and events), and ending with general properties broadly impacting multiple world elements (environments). Levels in the novelty hierarchy were intended to be mutually exclusive. Novelties of varying difficulty can arise from each level of the novelty hierarchy, though different novelty-handling techniques and more sophisticated world models may be required at higher numbered levels (Novelty Working Group, 2023). Note that the term "hierarchy" is used here for historical reasons; no hierarchical relationship is required or expected between these novelty characterizations.

| Tier | Category | Definition | Notes |
|---|---|---|---|
| Single Entities | 1 – Objects | An entity in the environment that is not the point-of-view agent and does not have goal-oriented behavior. | Objects may experience state changes as a result of actions by the i) point-of-view agent, ii) an external agent, iii) an event, iv) or a non-volitional behavior. |
| Single Entities | 2 – Agents | An entity in the environment that is not the point-of-view agent and has goal-oriented behavior. (People will disagree on what goal-oriented means, that is fine.) | Same as #1, except must have a goal. Goal does not have to be obvious. Random exploration can be a goal. |
| Single Entities | 3 – Actions | A goal-oriented behavior of an external agent that is not the point-of-view agent. | Goal does not have to be obvious. Random exploration can be a goal. Changes in movements and state-change attributes (a pendulum) of non-volitional entities (objects) are NOT considered actions. |
| Multiple Entities | 4 – Relations | Static property of the relationship between multiple entities. | Can be spatial, temporal, or other. Can include any combination of entities (objects and agents) and can include the point-of-view agent. |
| Multiple Entities | 5 – Interactions | Dynamic property of behaviors or actions that impacts multiple entities. | Can be spatial, temporal, or other. Can include any combination of entities (objects and agents) and can include the point-of-view agent. |
| Complex Phenomena | 6 – Environments | A change in an element of an open-world space that may impact the entire task space and is independent of a specific entity. | Can be general across the entire task space (e.g., global temperature change) or varied in space and time (e.g., partial or temporary darkness or wind). |
| Complex Phenomena | 7 – Goals | A change in the purpose of goal-oriented behavior by an agent in the environment that is not the point-of-view agent. | If a domain already has team members of the AI agent (embodied or non-embodied) that alter goals, then changes in team or AI agent goal (directives) communicated by this external team member are in-scope. |
| Complex Phenomena | 8 - Events | A state change or series of state changes that are not each the result of volitional action by an agent. | This includes state changes with specific preconditions. |

*Table 2. Novelty Hierarchy, ca. January 2022 (Kildebeck et al, 2022)*

The formalizations provided here show that the T-SAL R-Transformations and T-Transformations are sufficient for formalizing the novelty categorizations created for the SAIL-ON program. Future research should attempt to investigate what classes of novelty are difficult or easy for agents with various capabilities to reason about.

This section is organized as follows: there are eight novelty hierarchy levels. For each level, we first provide a natural language definition and clarifications shared in the DARPA SAIL-ON program (Kildebeck et al, 2022) to guide those groups responsible for generating novelty in these categories as to what novelties are admissible in each category. For each of the eight novelty hierarchy levels, we then provide a formal definition, first described in plain language then in L(T-SAL). The L(T-SAL) description of each level provides a decision function of the form *has<X>Novelty*(***t***, *d*), where "<X>" is the novelty type, ***t*** is a transformation sequence and *d* a T-SAL domain. This function decides whether the novelty described by ***t*** belongs to category X. Finally, for each level we provide one or more T-SAL-CR examples of the category as transformation sequences.

The ambiguity in the natural language definitions described below is such that novelties generated by various groups adhered to different standards (NWG, 2023). It is our hope that the formal standards presented here will help future researchers to adhere to common standards.

**1 – Objects**

*Natural language definition*: An entity in the environment that does not have goal-oriented behaviors.

*Clarifications*: Objects may experience state changes as a result of actions by i) the point-of-view agent, ii) an external agent, or iii) an event. This level is explicitly differentiated from "instance" novelty, which arises when new objects of an existing type are observed.

*Formal definition*: To introduce object novelty, a transformation sequence must either 1) add a new type that inherits from OBJECT, 2) add a new function that relates that object type to something else, or 3) change the initial distribution of such a function (via a new fluent generator). In addition, to ensure relevance, the new type or function must impact domain dynamics, through the preconditions of an action model, the conditions of a process model, and/or the triggers of an event model.

*hasObjectNovelty* (***t***, *d, sg*) ≡
 ∃*i, f, j, p, c, $d_N$*: 0 ≤ i < |***t***|
  ∧ $d_N$ = *applySequence*(***t***, *d*)
  ∧ (((*transformation-type*(***t****$_i$*) = ADDTYPE
    ∨ *transformation-type*(***t****$_i$*) = ADDTYPEPARENT
    ∨ *transformation-type*(***t****$_i$*) = REMOVETYPEPARENT)
   ∧ *p* = *transformation-argument*(***t****$_i$*, PARENT)
   ∧ *c* = *transformation-argument*(***t****$_i$*, CHILD)
   ∧ *f* ∈ *domain-functions*(*d*)

∧ *derived-from*(*c*, OBJECT, *d*$_N$)
                    ∧ *derived-from*(*p*, *function-argument-types*(*f*)$_j$, *d*)
                    ∧ ¬*derived-from*(*c*, *function-argument-types*(*f*)$_j$, *d*)
                    ∧ *relevant-function*(*function-name*(*f*), *d*))
            ∨ (*transformation-type*(***t***$_i$) = ADDFUNCTION ∧ *f* = *transformation-argument*(***t***$_i$, FUNCTION)
                    ∧ *derived-from*(*function-argument-types*(*f*)$_j$, OBJECT)
                    ∧ *relevant-function*(*function-name*(*f*), *d*$_N$))
            ∨ (*transformation-type*(***t***$_i$) = ADDFLUENTGENERATOR
                    ∧ *f* = *fluent-generator-function*(*transformation-argument*(***t***$_i$, FLUENTGENERATOR))
                    ∧ *derived-from*(*function-argument-types*(*f*)$_j$, OBJECT)
                    ∧ *relevant-function*(*function-name*(*f*), *d*$_N$))

*Required Definitions:*

The relation *derived-from* is defined in Appendix A.

*relevant-function*(*fn*, *d*) ≡
    (∃*cnd*: ((∃*a* ∈ *domain-actions*(*d*) ∧ *cnd* ∈ *action-preconditions*(*a*))
            ∨ (∃*ev* ∈ *domain-events*(*d*) ∧ *cnd* ∈ *event-triggers*(*ev*)) ∨
            ∨ (∃*p* ∈ *domain-processes*(*d*) ∧ *cnd* ∈ *process-conditions*(*ev*))
        ∧ (*function-in-condition*(*fn*, *cnd*))
            ∨ ∃*ax*: *function-affects-axiom*(*fn*, *ax*, *d*) ∧ *function-in-condition*(*fn*, *cnd*))
    ∨ *function-affects-performance*(*fn*, *d*)

*function-affects-performance*(*fn*, *d*) ≡
    *function-in-calculation*(*fn*, *domain-performance-function*(*d*))

*function-affects-axiom*(*fn*, *ax, d*) ≡
    *function-in-condition*(*fn*, *axiom-condition*(*ax*), *d*)
    ∨ ∃*ax*$_M$: *ax*$_M$ ∈ *domain-axioms*(*d*)
            ∧ *function-affects-axiom*(*fn*, *ax*$_M$, *d*)
            ∧ *function-in-condition*(*axiom-name*(*ax*$_M$), *axiom-condition*(*ax*), *d*)

*function-in-term*(*fn*, *te* ∈ *Te*) ≡ ⊥

*function-in-term*(*fn*, *te* = ⟨*f*, ***te***⟩ ∈ *F* × ***Te***) ≡
    *function-name*(*f*) = *fn*
    ∨ (∃*i*: 0≤ *i* < |***te***| ∧ *function-in-calculation*(*fn*, ***te***$_i$))

*function-in-condition*(*fn*, *cnd* ∈ *Te*) ≡
    *function-in-term*(*fn*, *cnd*)

*function-in-condition*(*fn*, ⟨*bop, cnd1, cnd2*⟩ ∈ *bOp* × *Cn* × *Cn*) ≡
    *function-in-condition*(*fn*, *cnd1*)
    ∨ *function-in-condition*(*fn*, *cnd2*)

*function-in-condition*(*fn*, ⟨FORALL, ***v***, *cnd1, cnd2*⟩) ≡
    *function-in-condition*(*fn*, *cnd1*)
    ∨ *function-in-condition*(*fn*, *cnd2*)

*function-in-condition*(*fn*, ⟨*ineq, calc1, calc2*⟩ ∈ *Ineq* × *Ca* × *Ca*) ≡
    *function-in-calculation*(*fn, calc1*)
    ∨ *function-in-calculation*(*fn, calc2*)

*function-in-calculation*(*fn, calc* ∈ *Te*) ≡
    *function-in-term*(*fn, calc*)

*function-in-calculation*(*fn*, ⟨*op, calc1, calc2*⟩ ∈ *Op* × *Ca* × *Ca*) ≡
    *function-in-calculation*(*fn, calc1*) ∨ *function-in-calculation*(*fn, calc2*)

*function-in-calculation*(*fn*, ⟨*aggop, v, con, calc*⟩ ∈ *AggOp* × *V* × *Cn* × *Ca*) ≡
    *function-in-condition*(*fn, con*) ∨ *function-in-calculation*(*fn, calc*)

*function-in-calculation*(*fn*, ⟨IF*, cnd, calc1, calc2*⟩ ∈ IF × *Cn* × *Ca* × *Ca*) ≡
    *function-in-condition*(*fn, cnd*) ∨ *function-in-calculation*(*fn, calc1*) ∨ *function-in-calculation*(*fn, calc2*)

*function-argument-types*(*f*) = [*function-argument-type*(*f, function-argument*(*f*)$_i$)]$_{i=1..|function-argument(f)|}$

*fluent-generator-function(fg, d)* = argmin$_f$ *f* ∈ *domain-functions*(*d*)
                                         ∧ *function-name*(*f*) = *fluent-generator-function-name*(*fg*)

*Example*: Every cart starts with a block within 2 distance units (3 carts standard):

    ADDVALUEGENERATOR(NEARBY-STARTS, FILTER(
                                    DRAWTUPLE([POSITION, POSITION], ["cart", "block"])),
                                    λ(x, y) → {|x – y| < 2})
    ADDVALUEGENERATOR(START-POSITIONS, NEWSET(NDRAWS(NEARBY-STARTS, 3))
    ADDVALUEGENERATOR(CLOSE-BLOCKS, NDRAWS(BLOCKGROUP, 3))
    ADDVALUEGENERATOR(FAR-BLOCKS, DIFFERENCE(BLOCKGROUP, CLOSE-BLOCKS))
    ADDFLUENTGENERATOR(CART-POSITION, ALLPERMUTATIONS([CARTGROUP], START-POSITION.CART)
    ADDFLUENTGENERATOR(BLOCK-POSITION,
    COMBINEFUNCTIONS(ALLPERMUTATIONS([CLOSE-BLOCKS], START-POSITION.BLOCK)
                    ALLPERMUTATIONS([FAR-BLOCKS], POSITION))

*Example*: Initial block velocities are doubled:
    *Original generator*:
        FLUENTGENERATOR(BLOCK-VELOCITY, ALLPERMUTATIONS([BLOCKGROUP], RANDOMINSET([1, -1]))
    ADDFLUENTGENERATOR(BLOCK-VELOCITY, ALLPERMUTATIONS([BLOCKGROUP], RANDOMINSET([2, -2]))

**2 – Agents**

*Natural language definition*: An entity in the environment that is not the point-of-view agent and has goal-oriented behavior. The agent is most frequently an external agent but can in some instances be the point-of-view agent. The goal of the agent does not have to be obvious.

*Clarifications*: Researchers will disagree on what 'goal-oriented behavior' means. This is fine, and researchers have flexibility within reason to determine best practices in specific domains. The goal of the agent does not have to be obvious. For example, random exploration could be a goal.

*Formal definition*: An agent-level novelty is defined identically to an object-level novelty, but using AGENT subtypes instead of OBJECT subtypes. Note that agents are primarily distinguished from objects in that every action has a performer which must be an agent.

*hasAgentNovelty* (***t**, d*) ≡
    ∃*i, f, j, p, c, $d_N$*: 0 ≤ i < |***t***|
        ∧ $d_N$ = *applySequence*(***t***, *d*)
        ∧ (((*transformation-type*(***t**_i*) = ADDTYPE
             ∨ *transformation-type*(***t**_i*) = ADDTYPEPARENT
             ∨ *transformation-type*(***t**_i*) = REMOVETYPEPARENT)
          ∧ *p* = *transformation-argument*(***t**_i*, PARENT)
          ∧ *c* = *transformation-argument*(***t**_i*, CHILD)
          ∧ *f* ∈ *domain-functions*(*d*)
          ∧ *derived-from*(*c*, AGENT, $d_N$)
          ∧ *derived-from*(*p*, *function-argument-types*(*f*)_*j*, *d*)
          ∧ ¬*derived-from*(*c*, *function-argument-types*(*f*)_*j*, *d*)
          ∧ *relevant-function*(*function-name*(*f*), *d*))
        ∨ (*transformation-type*(***t**_i*) = ADDFUNCTION ∧ *f* = *transformation-argument*(***t**_i*, FUNCTION)
            ∧ *derived-from*(*function-argument-types*(*f*)_*j*, AGENT)
            ∧ *relevant-function*(*function-name*(*f*), $d_N$))
        ∨ (*transformation-type*(***t**_i*) = ADDFLUENTGENERATOR
            ∧ *f* = *fluent-generator-function*(*transformation-argument*(***t**_i*, FLUENTGENERATOR))
            ∧ *derived-from*(*function-argument-types*(*f*)_*j*, AGENT)
            ∧ *relevant-function*(*function-name*(*f*), $d_N$))

*Required Definitions:*

The relation *derived-from* is defined in Appendix A.

*Example*: A transformation sequence changes the amount of force other agents use when pushing, referenced by the push action (see Section 3.1).

    ADDVALUEGENERATOR(OTHER-AGENTS, NEWSET(DIFFERENCE(AGENTS, [AGENT1])))
    ADDFLUENTGENERATOR(AGENT-FORCE, ALLPERMUTATIONS([OTHER-AGENTS],
                                      UNIFORMDISTRIBUTION(0.5, 1.5)))

## 3 – Actions

*Natural language definition*: A goal-oriented behavior of an external agent that is not the point-of-view agent.

*Clarifications*: Non-goal-oriented movements of non-volitional objects are not considered actions.

*Formal definition*: For a transformation sequence to exhibit action novelty, the preconditions or effects of an action model must be changed. In addition, we require that the performer of the changed action cannot be of the same type $ty_{AG}$ as the point-of-view agent.

*hasActionNovelty* (***t*** = [***t**_0* … ***t**_n*], $ty_{AG}$, *d*) =
    ∃*i* < |***t***|, $d_p, d_T, a_T, n, v_T$, θ, ***v**_{BOUND}*:
        $d_p$ = *applySequence*([***t**_0* … ***t**_{i-1}*], *d*)
        ∧ ∃*a* ∈ *domain-actions*($d_p$): *n* = *action-name*(*a*)
            ∧ *transformation-argument*(***t**_i*, ACTIONNAME) = *n*

$$\land \textit{transformation-type}(\textbf{\textit{t}}_i) \in \{\text{ADDPRECONDITION, REMOVEPRECONDITION,}$$
$$\text{ADDACTIONEFFECT, REMOVEACTIONEFFECT}\}$$

$\land d_T = \textit{applySequence}(\textbf{\textit{t}}, d)$

$\land a_T \in \textit{domain-actions}(d_T) \land \textit{action-name}(a_T) = n$

$\land \textit{action-performer}(a_T) = v_T$

$\land \textbf{\textit{v}}_{BOUND} = \textit{bound-variables-in-conditions}(\textit{action-preconditions}(a_T), \textbf{\textit{v}}_{BOUND})$
$\cup (\{\textit{action-performer}(a_T)\} \cap V) \cup (\textit{action-parameters}(a_T)$

$\land \neg \exists \theta: (\textit{assignable-to}(ty_{AG}, v_T, \theta, d_T)$
$\land \forall cnd \in \textit{action-preconditions}(a_T): \textit{legal-condition}(cnd, \theta, \textbf{\textit{v}}_{BOUND}, d))$

*Required Definitions:*

The relation *bound-variables-in-conditions* and *assignable-to* are defined in Appendix A.

*Example*: An environment transformation adds an action that lets agents "donate" cart velocity to blocks

ADDACTION(DONATE, ?AG, [?C], [CART])
ADDPRECONDITION(DONATE, (= (CONTROLS ?AG ?C) TRUE))
ADDPRECONDITION(DONATE, (= (CART-VELOCITY ?C) ?V))
ADDACTIONEFFECT(DONATE, (SET (CART-VELOCITY ?C) 0))
ADDACTIONEFFECT (DONATE, (INCREASE (BLOCK-VELOCITY ?B) ?V))

*Example*: Cart pushes have a budget

ADDFUNCTION(PUSH-BUDGET)
ADDPRECONDITION(PUSH, (> (PUSH-BUDGET ?AG) ?FORCE))
ADDACTIONEFFECT (PUSH, (DECREASE (PUSH-BUDGET ?AG) ?FORCE))
ADDFLUENTGENERATOR(PUSH-BUDGET, ALLPERMUTATIONS([AGENTS], CONSTANTFUNCTION(100)))

## 4 – Relations

*Natural language definition*: Static properties of the relationships between multiple entities.

*Clarifications*: Can be spatial, temporal, or other. Can include the point-of-view agent and other entities.

*Formal definition*: A transformation sequence exhibits relation novelty if a static function (one not present in any effects or process changes) is created, has its distribution changed, or is added to triggers, conditions, or preconditions. The function must be relevant and must involve 2 or more entities (here, entities are specifically objects of types AGENT, OBJECT, or their subtypes).

*hasRelationNovelty*(**t**, d) =
 $\exists f, \textit{name}, i: 0 \leq i < |\textbf{\textit{t}}|$
  $\land ((\textit{transformation-type}(\textbf{\textit{t}}_i) = \text{ADDFUNCTION}$
   $\land \textit{function-name}(\textit{transformation-argument}(\textbf{\textit{t}}_i, \text{FUNCTION})) = \textit{name})$
   $\lor (\textit{transformation-type}(\textbf{\textit{t}}_i) \in \{\text{ADDPRECONDITION, REMOVEPRECONDITION}\}$
   $\land \textit{function-in-condition}(\textit{name}, \textit{transformation-argument}(\textbf{\textit{t}}_i, \text{PRECONDITION})))$
   $\lor (\textit{transformation-type}(\textbf{\textit{t}}_i) \in \{\text{ADDTRIGGER, REMOVETRIGGER}\}$
   $\land \textit{function-in-condition}(\textit{name}, \textit{transformation-argument}(\textbf{\textit{t}}_i, \text{TRIGGER})))$
   $\lor (\textit{transformation-type}(\textbf{\textit{t}}_i) \in \{\text{ADDPROCESSCONDITION, REMOVEPROCESSCONDITION}\}$

$\quad\quad\quad\quad\quad\quad\quad \land$ *function-in-condition(name, transformation-argument($t_i$, CONDITION))))*
$\quad\quad\quad\quad\quad \lor$ (*transformation-type*($t_i$) = ADDFLUENTGENERATOR
$\quad\quad\quad\quad\quad\quad \land$ *name = fluent-generator-function-name(*
$\quad\quad\quad\quad\quad\quad\quad\quad\quad\quad\quad\quad$ *transformation-argument($t_i$,* FLUENTGENERATOR))))
$\quad\quad\quad \land d_T$ = *applySequence*($t, d$)
$\quad\quad\quad \land f \in$ *domain-functions*($d_T$)
$\quad\quad\quad \land$ *function-name*($f$) = *name*
$\quad\quad\quad \land$ |*entities-in-function*($f$)| > 1
$\quad\quad\quad \land (\forall a \in$ *domain-actions*($d_T$) : $\nexists e \in$ *action-effects*($a$) : *effect-name*($e$) = *name*
$\quad\quad\quad\quad\quad \lor \forall ev \in$ *domain-events*($d_T$) : $\nexists e \in$ *event-effects*($ev$) : *effect-name*($e$) = *name*
$\quad\quad\quad\quad\quad \lor \forall p \in$ *domain-processes*($d_T$) : $\nexists c \in$ *process-changes*($p$) : *change-name*($c$) = *name*)

*Required Definitions:*

*entities-in-function*($f$) =
$\quad$ {$v \in$ *function-arguments* ($f$) | *derived-from*(*function-argument-type*($f,v$), OBJECT)
$\quad\quad\quad\quad\quad\quad\quad\quad\quad \lor$ *derived-from*(*function-argument-type*($f,v$), AGENT)}

*Example*: A transformation sequence asserts a minimum distance between every block and cart. This adds a new binary static function, MIN-DIST. The revised scenario generator assigns a random value between 0 and 1 to this function for each pair of block and cart. The PUSH action preconditions change to enforce this constraint.
$\quad\quad\quad$ ADDFUNCTION(⟨MIN-DIST, [?B, ?C], [BLOCK, CART], REAL⟩))
$\quad\quad\quad$ ADDPRECONDITION(PUSH, (= (CART-POSITION ?C) ?P))
$\quad\quad\quad$ ADDPRECONDITION(PUSH, (FORALL ?B (IS-BLOCK ?B)
$\quad\quad\quad\quad\quad\quad\quad\quad\quad\quad\quad\quad\quad\quad$ (AND (< (- (+ ?P ?FORCE) (BLOCK-POSITION ?B)) (MIN-DIST ?B ?C))
$\quad\quad\quad\quad\quad\quad\quad\quad\quad\quad\quad\quad\quad\quad\quad$ (< (- (BLOCK-POSITION ?B) (+ ?P ?FORCE)) (MIN-DIST ?B ?C))))
$\quad\quad\quad$ ADDFLUENTGENERATOR(MIN-DIST, ALLPERMUTATIONS([BLOCKS, CARTS], UNIFORMDISTRIBUTION(0, 1)))

**5 – Interactions**

*Natural language definition*: Dynamic property of behaviors or actions that impacts multiple entities.

*Clarifications*: Can be spatial, temporal, or other. Can include the point-of-view agent and other entities.

*Formal definition*: A transformation sequence exhibits interaction novelty if a fluent is added to or changed in the effects of an action. The fluent must name a relevant function that involves 2 or more entities (here, entities are specifically objects of types AGENT, OBJECT, or their subtypes).

*hasInteractionNovelty*($t, d$) =
$\quad \exists f, fname, aname, i: 0 \leq i < |t|$
$\quad\quad\quad\quad \land$ ((*transformation-type*($t_i$) ∈ {ADDACTIONEFFECT, REMOVEACTIONEFFECT}
$\quad\quad\quad\quad\quad\quad \land$ *effect-name(transformation-argument*($t_i$, EFFECT)) = *fname*)
$\quad\quad\quad\quad \land \exists a \in$ *domain-actions*($d$):
$\quad\quad\quad\quad\quad\quad$ *aname* = *action-name*($a$)
$\quad\quad\quad\quad\quad\quad \land$ *transformation-argument*($t_i$, ACTIONNAME) = *aname*

    ∧ *function-name*(*f*) = *fname*
    ∧ |*entities-in-function*(*f*)| > 1
    ∧ $d_T$ = *applySequence*(***t***, *d*)
    ∧ (∃$a_T$ ∈ *domain-actions*($d_T$) :
      *action-name*($a_T$) = *aname*
      ∧ ∃*e* ∈ *action-effects*($a_T$) : *effect-name*(*e*) = *fname*)

*Example*: A new action allows an agent to move to a nearby cart; this affects the binary function CONTROLS, which becomes dynamic and affects the relationship between agent and cart.
  ADDACTION(SWITCH-CARTS, [?C1, ?C2], [CART, CART])
  ADDPRECONDITION(SWITCH-CARTS, (= (CONTROLS ?AG ?C1) TRUE))
  ADDPRECONDITION(SWITCH-CARTS, (= (CONTROLS ?AG ?C2) FALSE))
  ADDPRECONDITION(SWITCH-CARTS, (= (CART-POSITION ?C1) ?P1))
  ADDPRECONDITION(SWITCH-CARTS, (= (CART-POSITION ?C2) ?P2))
  ADDPRECONDITION(SWITCH-CARTS, (< (- ?P2 ?P1) 1))
  ADDPRECONDITION(SWITCH-CARTS, (< (- ?P1 ?P2) 1))
  ADDACTIONEFFECT(SWITCH-CARTS, (= (CONTROLS ?AG ?C2) TRUE))
  ADDACTIONEFFECT(SWITCH-CARTS, (= (CONTROLS ?AG ?C1) FALSE))

## 6 – Environment

*Natural language definition*: A change in an element of an open-world space that may impact the entire task space and is independent of a specific entity.

*Clarifications*: Environment novelties may impact the entire task space identically (e.g., changing the temperature from a static value to a new static value) or may impact regions of the open world domain differentially (e.g., wind may be present in only part of a space and the intensity may ebb and flow over time).

*Formal definition*: An environment novelty is a transformation sequence in which environmental functions, processes, or events are modified. Environmental functions are those functions outside of agent control, meaning they are not affected by functions. Environmental processes and events are those that are conditioned on or triggered by only environmental functions and position-valued functions.

*hasEnvironmentNovelty*(***t***, *d*) =
 (∃*i*, $d_T$:
  0 ≤ *i* < |***t***|
  ∧ $d_T$ = *applySequence*(***t***, *d*)
  ∧ (*transformation-type*($t_i$) = ADDFLUENTGENERATOR
    ∧ ∃*f*, *fname*:
     *isEnvironmentalFunction*(*f*, $d_T$)
     ∧ *fname* = *function-name*(*f*)
     ∧ *transformation-argument*($t_i$, NAME) = *fname*)
   ∨ (*transformation-type*($t_i$) ∈ {ADDTRIGGER, REMOVETRIGGER, ADDEVENTEFFECT, REMOVEEVENTEFFECT,
      CHANGEPROBABILITY, CHANGEFREQUENCY}

$\quad\quad\quad\land \exists ev \in$ *domain-events(d_T)*:
$\quad\quad\quad\quad$ *event-name*($ev$) = *transformation-argument*($t_i$, EVENTNAME))
$\quad\quad\quad\quad\land$ *isEnvironmentalEvent*($ev$, $d_T$))
$\quad\quad\lor$ (*transformation-type*($t_i$) $\in$ {ADDPROCESSCONDITION, REMOVEPROCESSCONDITION,
$\quad\quad\quad\quad\quad\quad$ ADDPROCESSCHANGE, REMOVEPROCESSCHANGE}
$\quad\quad\land \exists p \in$ *domain-processes(d_T)*:
$\quad\quad\quad$ *process-name*($p$) = *transformation-argument*($t_i$, PROCESSNAME))
$\quad\quad\quad\land$ *isEnvironmentalProcess*($p$, $d_T$))

*Required definitions*:

*isEnvironmentalFunction*($f$, $d$) =
$\quad f \in$ *domain-functions*($d$)
$\quad\land \neg(\exists a, e: a \in$ *domain-actions*($d$) $\land e \in$ *action-effects*($a$) $\land$ *effect-name*($e$) = *function-name*($f$)
$\quad\land$ |*entities-in-function*($f$)| = 0

*isEnvironmentalProcess*($p$, $d$) =
$\quad \exists cnd \in$ *process-preconditions*($ev$), $f \in$ *domain-functions*($d$):
$\quad\quad$ *function-in-condition*(*function-name(f)*, *cnd*) $\land$ *isEnvironmentalFunction*($d$, $f$)
$\quad \forall cnd \in$ *process-preconditions*($p$), $f \in$ *domain-functions*($d$):
$\quad\quad \neg$ *function-in-condition*(*function-name(f)*, *cnd*)
$\quad\quad\quad \lor$ *isEnvironmentalFunction*($d$, $f$) $\lor$ *function-value-type*($f$) = POSITION

*isEnvironmentalEvent*($ev$, $d$) =
$\quad \exists cnd \in$ *event-triggers*($ev$), $f \in$ *domain-functions*($d$):
$\quad\quad$ *function-in-condition*(*function-name(f)*, *cnd*) $\land$ *isEnvironmentalFunction*($d$, $f$)
$\quad \forall cnd \in$ *event-triggers*($ev$), $f \in$ *domain-functions*($d$):
$\quad\quad \neg$ *function-in-condition*(*function-name(f)*, *cnd*)
$\quad\quad\quad \lor$ *isEnvironmentalFunction*($d$, $f$) $\lor$ *function-value-type*($f$) = POSITION

*Example*: A transformation introduces a "gravitational jump" that distorts pole angles every 20 time units. This is modeled by an environmental event, JUMP-PULLS.
$\quad$ ADDFUNCTION(JUMP-TIME, [], [])
$\quad$ ADDPROCESS(JUMP-TICKS, [], [])
$\quad$ ADDPROCESSCHANGE(JUMP-TICKS, (INCREASE (JUMP-TIME) (* DT 1)))
$\quad$ ADDEVENT(JUMP-PULLS, [?C], [CART])
$\quad$ ADDTRIGGER(JUMP-PULLS, (> (JUMP-TIME) 20))
$\quad$ ADDEVENTEFFECT(JUMP-PULLS, (INCREASE (POLE-ANGLE ?C) 5))
$\quad$ ADDEVENTEFFECT(JUMP-PULLS, (SET (JUMP-TIME) 0))
$\quad$ ADDFLUENTGENERATOR(JUMP-TIME, ALLPERMUTATIONS([], UNIFORMRANDOM(0, 20)))

**7 – Goals**

*Natural language definition*:  The purpose of a behavior by an agent in the environment.

*Clarifications*: Goal novelties resulting from environmental transformations are primarily considered to be goal changes of external agents in the environment that are not the point-of-view agent. Some domains may include teammates where the task goal is dynamic and communicated to the point-of-view agent as a directive or change in observable reward. In such domains, changes in the goals of the team or point-of-view agent communicated by external team members can be considered in-scope, but not required for point-of-view.

*Formal definition*: Goal novelties are those that include a T-Transformation changing how the performance function of a scenario generator is calculated.

*hasGoalNovelty*(*t, d*) ≡
 $\exists t \in \mathbf{t}$: *transformation-type*(*t*) = REPLACEPERFORMANCECALCULATION

*Example*: An environment transformation introduces a new performance calculation that rewards an agent lasting without winning or losing.

> ADDFUNCTION(CLOCK-TIME, [], [])
> ADDPROCESS(CLOCK-TICKS, [], [])
> ADDPROCESSCHANGE(CLOCK-TICKS, (INCREASE (CLOCK-TIME) (* DT 1)))
> REPLACEPERFORMANCECALCULATION((CLOCK-TIME))
> ADDDEFAULT(CLOCK-TIME, 0)

**8 – Events**

*Natural language definition*:  A state change or series of state changes that are not the result of volitional action by an external agent or the point-of-view agent.
*Clarifications*: Events include state changes with specific preconditions.

*Formal definition*: An event novelty is a transformation sequence that affects T-SAL events that have both environmental and non-environmental triggers. At least one trigger of a modified event must be based on an environment function and at least one must be directly affected by an action. Event novelty sequences include those that modify the distribution of an environmental function that triggers such an event.

*hasEventNovelty*(*t, d*) ≡
 $\exists ev, d_T, i$:
  *ev* ∈ *domain-events*(*d_T*)
  ∧ *d_T* = *applySequence*(*t*, *d*)
  ∧ ¬*isEnvironmentalEvent*(*ev*, *d_T*)
  ∧ 0 ≤ *i* < |*t*|
  ∧ ((*transformation-type*(*t_i*) ∈ {ADDTRIGGER, REMOVETRIGGER, ADDEVENTEFFECT, REMOVEEVENTEFFECT,
             CHANGEPROBABILITY, CHANGEFREQUENCY}
    ∧ *event-name*(*ev*) = *transformation-argument*(*t_i*, EVENTNAME))
   ∨ (*transformation-type*(*t_i*) = ADDFLUENTGENERATOR
     ∧ $\exists f$: *isEnvironmentalFunction*(*f*, *d_T*)

∧ *transformation-type*(**t**$_i$, NAME) = *function-name*(*f*)
∧ ∃*cnd* ∈ *event-triggers*(*ev*): *function-in-condition*(*function-name*(*f*), *cnd*))

*Example*: An environment transformation introduces a "gravitational anomaly" that pulls nearby carts and blocks based on their position, if they are moving. This is modelled by a non-environmental event, GRAVITY-PULLS, with a varying position trigger, an environmental function trigger, and a third trigger (velocity) under the agent's control.

>ADDFUNCTION(GRAVITY-LOCATION)
>ADDPROCESS(GRAVITY-WAVE-MOVES, [], [])
>ADDPROCESSCHANGE(GRAVITY-WAVE-MOVES, (INCREASE (GRAVITY-LOCATION) (* DT 1)))
>ADDEVENT(GRAVITY-WAVE-RESETS, [], [])
>ADDTRIGGER(GRAVITY-WAVE-RESETS, (>= (GRAVITY-LOCATION) 20))
>ADDEVENTEFFECT(GRAVITY-WAVE-RESETS, (SET (GRAVITY-LOCATION) -20))
>ADDEVENT(GRAVITY-PULLS, [?C], [CART])
>CHANGEEVENTFREQUENCY(GRAVITY-PULLS, 4)
>ADDTRIGGER(GRAVITY-PULLS, (> (CART-VELOCITY ?C) .01)))
>ADDTRIGGER(GRAVITY-PULLS, (= (CART-POSITION ?C) ?P)))
>ADDTRIGGER(GRAVITY-PULLS, (= (GRAVITY-LOCATION) ?WAVE-POS)))
>ADDEVENTEFFECT(GRAVITY-PULLS, (INCREASE (CART-VELOCITY ?C) (* DT (LOG-DISTANCE ?P ?WAVE-POS))))
>ADDFLUENTGENERATOR(GRAVITY-LOCATION, ALLPERMUTATIONS([], UNIFORMRANDOM(-20, 20)))

# 5 Discussion and Future Work

In this work, we have set up and formally defined a framework for describing a general class of environments and environment transformations. This is a necessary precursor to general agreement on definitions and measurement of the robustness of artificially intelligent agents. These definitions will be useful to future studies of general agents that take on very difficult situations flexibly and adaptively without human assistance. The "novelty hierarchy" is a first step in exploring possible classes of transformation sequences. Future work will use the formalism described here to define more useful assumptions to guide research.

Human design choices about T-SAL environments may have significant impact on how a transformation is described. In some cases, this has been observed to change category membership for the novelty hierarchy described here. This ambiguity is exacerbated when considering agents that have learned their own environment descriptions. To such an agent, the original environment may have a different definition than to a describing human – even if it's consistent with the same observations. Therefore, a novelty hierarchy categorization might be different than to an observing human.

A known failing of T-SAL-CR is that it does not describe observation of the environment; clearly, the environment is responsible for providing an agent affordances in observation, which can change over time. For example, a robot may gain or lose awareness of its location via a GPS sensor. T-SAL-CR cannot describe what aspects of the state are observable, and therefore the T-SAL transformation language also cannot describe changes in observability. Future work should remedy this problem.

## Acknowledgements

This material is based upon work supported by the Defense Advanced Research Projects Agency (DARPA) under Contract No. HR001121C0236. Any opinions, findings and conclusions or recommendations expressed in this material are those of the author(s) and do not necessarily reflect the views of DARPA.

# Appendix A  T-SAL Environment Legality

The definitions below describe conditions for a legal T-SAL environment.

The relation *derived-from*: *Ty × Ty × D* indicates whether two types have a derivation relationship; i.e., there is a path of supertypes in the domain from the proposed descendant type to the ancestor type. All top-level types that do not derive from other basic types derive from object.

*derived-from*($ty_D$, $ty_A$, *d*) ≡
  ⟨$ty_D$, $ty_A$⟩ ∈ *domain-supertypes(d)*
  ∨ ∃*ty*: ⟨$ty_D$, $ty_A$⟩ ∈ *domain-supertypes(d)* ∧ *derived-from(ty, $ty_A$, d)*

*derived-from*($ty_D$, OBJECT, *d*) ≡
  ∀*ty* ∈ *domain-types(d)*: ⟨$ty_D$, *ty*⟩ ∉ *domain-supertypes(d)*
    ∧ $ty_D$ ∉ {OBJECT, REAL, INTEGER, BOOLEAN, AGENT}

To describe variable-type relationships, we use sets of ⟨variable, type⟩ pairs that allow mapping of a variables to an assigned type. We denote such a set θ = (⟨$v_0$, $ty_0$⟩, ⟨$v_1$, $ty_1$⟩, … ⟨$v_n$, $ty_n$⟩). The function *variable-type*: θ × *V* → (*S* ∪ {⊥})  gives the type of a variable found in or ⊥ if a variable is not found in the set; *variable-undefined* is true iff a variable is not Found in the set.

*variable-type*(θ = (⟨$v_0$, $ty_0$⟩, ⟨$v_1$, $ty_1$⟩, … ⟨$v_n$, $ty_n$⟩), *v*) =
  *if* (|θ| = 0, TRUE: ⊥,
          FALSE: *if* ($v_0$ = *v*, TRUE: $ty_0$,
                          FALSE: *variable-type*((⟨$v_1$, $ty_1$⟩, … ⟨$v_n$, $ty_n$⟩), *v*)

*variable-undefined*(θ = (⟨$v_0$, $ty_0$⟩, ⟨$v_1$, $ty_1$⟩, … ⟨$v_n$, $ty_n$⟩), *v*) ≡ ¬∃*i*: $v_i$ = *v*

The relation *assignable-to*: *Ty × Ca ×* θ *× D* indicates whether a particular calculation is assignable to a particular type, based on existing variable types and domain supertype information.

*assignable-to*(*ty, calc* ∈ *V,* θ*, d*) ≡
  ¬*variable-undefined*(θ*, v*) ∧ *derived-from*(*variable-type*(θ*, calc*)*, ty, d*)

*assignable-to*(*ty, o* ∈ *O,* θ*, d*) ≡ ⟨*o, $ty_s$*⟩ ∈ *domain-constants(o)* ∧ *derived-from*($ty_s$*, ty, d*)

*assignable-to*(*ty, i* ∈ $\mathbb{I}$*,* θ*, d*) ≡ *derived-from*(*ty,* INTEGER*, d*)

*assignable-to*(*ty, r* ∈ $\mathbb{R}$*,* θ*, d*) ≡ *derived-from*(*ty,* REAL*, d*)

*assignable-to*(*ty, b* ∈ {TRUE, FALSE}*,* θ*, d*) ≡ *derived-from*(*ty,* BOOLEAN*, d*)

*assignable-to*(*ty,* ⟨*f,* **te**⟩ ∈ *F ×* **Te***,* θ*, d*) ≡ *derived-from*(*function-value-type(f), ty, d*)

*assignable-to*(*ty,* ⟨*op, calc1, calc2*⟩ ∈  *Op × Ca × Ca,* θ*, d*) ≡
  *derived-from*(*ty,* REAL*, d*)
  ∨ (*derived-from*(*ty,* INTEGER*, d*)
    ∧ (*op* = :UNIFORM
      ∨ (*op* ∈ (+, -, *, /) ∧ *assignable-to*(INTEGER*, calc1,* θ*, d*) ∧ *assignable-to*(INTEGER*, calc2,* θ*, d*))))

*assignable-to*(*ty*, ⟨*aggop, v, con, calc*⟩ ∈ *AggOp* × *V* × *Cn* × *Ca*, θ, *d*) ≡
  ∃θ$_N$ ⊃ θ: ¬*variable-undefined*(θ$_N$, *v*)
      ∧ (*derived-from*(*ty*, REAL, *d*) ∧ *assignable-to*(*calc*, REAL, θ, *d*)
        ∨ *derived-from*(*ty*, INTEGER, *d*) ∧ *assignable-to*(*calc*, INTEGER, θ, *d*))

*assignable-to*(*ty*, ⟨*cnd, calc$_T$, calc$_F$*⟩ ∈ IF *Cn* × *Ca* × *Ca*, θ, *d*) ≡
  *derived-from*(*ty*, REAL, *d*) ∧ *assignable-to*(*calc$_T$*, REAL, θ, *d*) ∧ *assignable-to*(*calc$_F$*, REAL, θ, *d*)
  ∨ *derived-from*(*ty*, INTEGER, *d*) ∧ *assignable-to*(*calc$_T$*, INTEGER, θ, *d*) ∧ *assignable-to*(*calc$_F$*, INTEGER, θ, *d*)

All variables and values are legal terms. A function term is legal if the terms used with it are terms assignable to that term's argument types.

*legal-term*(*te* ∈ *V* ∪ *Va*, θ, ***v*$_{BOUND}$**, *d*) ≡ TRUE

*legal-term*(*te* = ⟨*fn*, ***te***⟩ ∈ *Fn* × ***Te***, θ, ***v*$_{BOUND}$**, *d*) ≡
  ∃*f* ∈ *domain-functions*(*d*):
    *function-name*(*f*) = *fn*
    ∧ |*function-arguments*(*f*)| = |***te***|
    ∧ ∀*i*, 0 ≤ *i* < |***te***| : *assignable-to*(*function-argument-type*(*f*, *function-arguments*(*f*)$_i$), ***te****$_i$*, θ, *d*)
        ∧ *legal-calculation*(*ty*, ***te****$_i$*, θ, *d*)

A calculation is legal if it's a bound term, bound fluent with correct argument types, or an operation with correct argument types. An aggregation is legal if its constraint condition and subcalculation are legal, and the calculation matches the supertype.

*legal-calculation*(*ty*, *te* ∈ *Te*, θ, *d*) ≡
  *assignable-to*(*ty*, *te*, θ, *d*)
  ∧ (*te* ∉ *V* ∧ ¬*variable-undefined*(θ, *te*))

*legal-calculation*(*ty*, ⟨*op, calc1, calc2*⟩ ∈ *Op* × *Ca* × *Ca*, θ, *d*) ≡
  *assignable-to*(*ty*, ⟨*op, calc1, calc2*⟩, θ, *d*)
  ∧ ((*op* ≠ :UNIFORM ∧ *assignable-to*(REAL, *calc1*, θ, *d*) ∧ *assignable-to*(REAL, *calc2*, θ, *d*))
    ∨ (*op* = :UNIFORM ∧ *assignable-to*(INTEGER, *calc1*, θ, *d*) ∧ *assignable-to*(INTEGER, *calc2*, θ, *d*))

*legal-calculation*(*ty*, ⟨*aggop, v, con, calc*⟩ ∈ *AggOp* × *V* × *Cn* × *Ca*, θ, *d*) ≡
  *assignable-to*(*ty*, ⟨*aggop, v, con, calc*⟩, θ, *d*)
  ∧ ∃θ$_N$: *extends*(θ$_N$, θ)
      *variable-undefined*(θ, *v*) ∧ ¬*variable-undefined*(θ$_N$, *v*)
      ∧ *legal-condition*(*con*, θ$_N$, *d*)
      ∧ *legal-calculation*(*ty*, *calc*, θ$_N$, *d*)

*legal-calculation*(*ty*, ⟨IF, *cnd, calc$_T$, calc$_F$*⟩ ∈ IF *Cn* × *Ca* × *Ca*, θ, *d*) ≡
  *assignable-to*(*ty*, ⟨*cnd, calc$_T$, calc$_F$*⟩, θ, *d*)
  ∧ *legal-condition*(*cnd*, θ$_N$, *d*)
  ∧ *legal-calculation*(*ty*, *calc$_T$*, θ$_N$, *d*)
  ∧ *legal-calculation*(*ty*, *calc$_F$*, θ$_N$, *d*)

For a condition to be legal, all bindings must have constituent types throughout subconditions. An "and" or "or" condition is legal if its two condition arguments are legal, and a forall condition is legal if its two condition arguments are legal, and its variables are not previously defined. Comparisons with an inequality are legal only if all of the calculation variables are bound. In the case of a single variable being equal to a bound calculation, that variable becomes bound.

*legal-condition*(*te* ∈ *Te*, θ, ***v**$_{BOUND}$*, *d*) ≡
  *legal-term*(*te*, θ, ***v**$_{BOUND}$*, *d*) ∧ *assignable-to*(BOOLEAN, *te*, θ, *d*)

*legal-condition*(⟨*bop*, *cnd1*, *cnd2*⟩ ∈ *BOp* × *Cn* × *Cn*, θ, ***v**$_{BOUND}$*, *d*) ≡
  *legal-condition*(*cnd1*, θ, ***v**$_{BOUND}$*, *d*)
  ∧ *legal-condition*(*cnd2*, θ, ***v**$_{BOUND}$*, *d*)
  ∧ *bop* ∈ {AND, OR}

*legal-condition*(⟨FORALL, ***v***, *cnd1*, *cnd2*⟩, θ, ***v**$_{BOUND}$*, *d*) ≡
  ∃θ$_N$, ***v1**$_{INNER-BOUND}$*, ***v2**$_{INNER-BOUND}$*:
    *extends*(θ$_N$, θ)
    ∧ ∀*v* ∈ ***v***: *variable-undefined*(θ, *v*) ∧ ¬*variable-undefined*(θ$_N$, *v*)
    ∧ ***v1**$_{INNER-BOUND}$* = *condition-bound-variables*(*cnd1*, ***v1**$_{INNER-BOUND}$*) ∪ ***v**$_{BOUND}$* ∪ ***v***
    ∧ *legal-condition*(*cnd1*, θ$_N$, ***v1**$_{INNER-BOUND}$*, *d*)
    ∧ ***v2**$_{INNER-BOUND}$* = *condition-bound-variables*(*cnd2*, ***v2**$_{INNER-BOUND}$*) ∪ ***v1**$_{INNER-BOUND}$* ∪ ***v***
    ∧ *legal-condition*(*cnd2*, θ$_N$, ***v2**$_{INNER-BOUND}$*, *d*)

*legal-condition*(⟨*ineq*, *calc1*, *calc2*⟩ ∈ *Ineq* × *Ca* × *Ca*, θ, ***v**$_{BOUND}$*, *d*) ≡
  ∃*ty* ∈ *Ty*:
    *legal-calculation*(*ty*, *calc1*, *restrict*(θ, ***v**$_{BOUND}$*), *d*)
    ∧ *legal-calculation*(*ty*, *cnd2*, *restrict*(θ, ***v**$_{BOUND}$*), *d*)
    ∧ (*derived-from*(REAL, *ty*) ∨ *derived-from*(INTEGER, *ty*) ∨ *ineq* ∈ {=, ≠})
    ∧ (*ineq* ∈ {=} ∨ (*variables-in-calculation*(*calc1*) ∪ *variables-in-calculation*(*calc2*)) ⊂ ***v**$_{BOUND}$*))

*restrict*(θ = (⟨*v*$_0$, *ty*$_0$⟩, … ⟨*v*$_n$, *ty*$_n$⟩), ***v**$_{BOUND}$*) =
  *concat*(*if*(*v*$_0$ ∈ ***v**$_{BOUND}$*, TRUE: (⟨*v*$_0$, *ty*$_0$⟩), FALSE: ()),
    *restrict*((⟨*v*$_1$, *ty*$_1$⟩, … ⟨*v*$_n$, *ty*$_n$⟩), ***v**$_{BOUND}$*)

*extends*(θ$_B$ = (⟨*v*$_{B,0}$, *ty*$_{B,0}$⟩, … ⟨*v*$_{B,m}$, *ty*$_{B,m}$⟩), θ$_S$ = (⟨*v*$_{S,0}$, *ty*$_{S,0}$⟩, … ⟨*v*$_{S,m}$, *ty*$_{S,m}$⟩)) ≡
  *m* > *n* ∧ ∀i, 0 ≤ *i* ≤ n: *v*$_{B,i}$ = *v*$_{S,i}$ ∧ *ty*$_{B,i}$ = *ty*$_{S,i}$

*bound-variables-in-conditions*((*pc*$_0$… *pc*$_N$) ∈ ***PC***, ***v**$_{BOUND}$*) =
  ⋃$_i$ *condition-bound-variables*(*cnd*, *pc*$_i$)

*condition-bound-variables*(*te* ∈ *Te*, ***v**$_{BOUND}$*) = ∅

*condition-bound-variables*(⟨OR, *cnd1*, *cnd2*⟩ ∈ OR × *Cn* × *Cn*, *d*, ***v**$_{BOUND}$*) = ∅

*condition-bound-variables*(⟨AND, *cnd1*, *cnd2*⟩ ∈ AND × *Cn* × *Cn*, *d*, ***v**$_{BOUND}$*) =
  *condition-bound-variables*(*cnd1*, *d*, ***v**$_{BOUND}$*) ∪ *condition-bound-variables*(*cnd2*, *d*, ***v**$_{BOUND}$*)

*condition-bound-variables*(⟨FORALL, ***v***, *cnd1*, *cnd2*⟩ ∈ FORALL × ***V*** × *Cn* × *Cn*, *d*, ***v**$_{BOUND}$*) = ∅

*condition-bound-variables*(⟨*ineq, calc1, calc2*⟩ ∈ *Ineq* × *Ca* × *Ca*, *d*, $v_{BOUND}$) =
$$\begin{cases} calc1 \in V \land \textit{variables-in-calculation}(calc2) \subset v_{BOUND} & \{calc1\} \\ calc2 \in V \land \textit{variables-in-calculation}(calc1) \subset v_{BOUND} & \{calc2\} \\ \textit{otherwise} & \emptyset \end{cases}$$

*variables-in-term*(*te* ∈ *Va*) ≡ ∅

*variables-in-term*(*te* ∈ *V*) ≡ {*te*}

*variables-in-term*(*te* = ⟨*fn*, **te**⟩ ∈ *Fn* × **Te**) ≡ ⋃$_i$ *variables-in-term*(**te**$_i$)

*variables-in-calculation*(*calc* ∈ *Te*) ≡ ⋃$_i$ *variables-in-term*(**te**$_i$)

*variables-in-calculation*(*calc* = ⟨*op, calc1, calc2*⟩ ∈ *Op* × *Ca* × *Ca*) ≡
  *variables-in-calculation*(*calc1*) ∪ *variables-in-calculation*(*calc2*)

*variables-in-calculation*(*calc* = ⟨*aggop, v, con, calcsub*⟩ ∈ *AggOp* × *V* × *Cn* × *Ca*) ≡
  (*variables-in-calculation*(*calcsub*) ∪ *variables-in-condition*(*con*)) / {*v*}

*variables-in-calculation*(*calc* = ⟨IF, *cnd, calc1, calc2*⟩ ∈ IF × *Cn* × *Ca* × *Ca*) ≡
  *variables-in-condition*(*cnd*) ∪ *variables-in-calculation*(*calc1*) ∪ *variables-in-calculation*(*calc2*)

*variables-in-condition*(*te* ∈ *Te*) ≡ *variables-in-term*(*te*)

*variables-in-condition*(⟨*bop, cnd1, cnd2*⟩ ∈ *BOp* × *Cn* × *Cn*) ≡
  *variables-in-condition*(*cnd1*) ∪ *variables-in-condition*(*cnd2*))

*variables-in-condition*(⟨FORALL, *v, cnd1, cnd2*⟩ ∈ FORALL × *V* × *Cn* × *Cn*) ≡
  *variables-in-condition*(*cnd1*) ∪ *variables-in-condition*(*cnd2*)) / {*v*}

*variables-in-condition*(⟨*ineq, calc1, calc2*⟩ ∈ *Ineq* × *Ca* × *Ca*) ≡
  *variables-in-calculation*(*calc1*) ∪ *variables-in-calculation*(*calc2*))

*effect-arguments-bound*(*e, d*, $v_{BOUND}$) =
  (*effect-argument-list*(*e, d*) ∩ *V*) ⊂ $v_{BOUND}$

*effect-argument-list*(*e, d*) =
  {*te* | ∃*f, v*: *f* ∈ *domain-functions*(*d*) ∧ *function-name*(*f*) = *effect-name*(*e*)
       ∧ *v* ∈ *function-arguments*(*f*) ∧ *effect-argument*(*e, v*) = *te*}

*change-arguments-bound*(*c, d*, $v_{BOUND}$) =
  (*change-argument-list*(*c, d*) ∩ *V*) ⊂ $v_{BOUND}$

*change-argument-list*(*c, d*) =
  {*te* | ∃*f, v*: *f* ∈ *domain-functions*(*d*) ∧ *function-name*(*f*) = *change-name*(*c*)
       ∧ *v* ∈ *function-arguments*(*f*) ∧ *change-argument*(*c, v*) = *te*}

*axiom-argument-list*(*ax, d*) =
  {*te* | ∃*f, v*: *f* ∈ *domain-functions*(*d*) ∧ *function-name*(*f*) = *axiom-name*(*ax*)
       ∧ *v* ∈ *function-arguments*(*f*) ∧ *axiom-argument*(*ax, v*) = *te*}

Effects are legal if and only if they match a fluent properly, and match necessary variable type bindings θ.

   *legal-effect*(*e*, θ, $v_{BOUND}$, *d*) ≡
     ∃*f* ∈ *domain-functions*(*d*):
      *function-name*(*f*) = *effect-name*(*e*)
      ∧ (*derived-from*(REAL, *function-value-type*(*f*)) ∨ *derived-from*(INTEGER, *function-value-type*(*f*))
        ∨ *effect-modification*(*e*) = SET)
      ∧ *assignable-to*(*function-value-type*(*f*), *effect-value*(*e*), θ, *d*)
      ∧ ({*effect-value*(*e*)} ∩ V) ⊆ $v_{BOUND}$
      ∧ *effect-arguments-bound*(*e*, $v_{BOUND}$, *d*)
      ∧ ∀*v* ∈ *function-arguments*(*f*): *assignable-to*(*effect-argument*(*e*, *v*), *function-argument-type*(*f*, *v*), θ, *d*)

Continuous changes are legal if and only if they match a function properly, as well as variable types θ.

   *legal-change*(*c*, θ, *d*) ≡
     ∃*f* ∈ *domain-functions*(*d*):
      *function-name*(*f*) = *change-name*(*c*)
      ∧ (*derived-from*(REAL, *function-value-type*(*f*)) ∨ *derived-from*(INTEGER, *function-value-type*(*f*)))
      ∧ *legal-calculation*(*function-value-type*(*f*), *change-value*(*c*), θ, *d*)
      ∧ *change-arguments-bound*(*c*, $v_{BOUND}$, *d*)
      ∧ ∀*v* ∈ *function-arguments*(*f*): *assignable-to*(*change-argument*(*c*, *v*), *function-argument-type*(*f*, *v*), θ, *d*)

   *legal-function*(*f*, *d*) ≡
     ∀*v* ∈ *function-arguments*(*f*): *function-argument-type*(*f*, *v*) ∈ *domain-types*(*d*)
     ∧ *function-value-type*(*f*) ∈ *domain-types*(*d*)

   *legal-axiom*(*ax*, θ, *d*) ≡
     ∃θ, *f*, $v_{BOUND}$:
      *f* ∈ *domain-functions*(*d*)
      ∧ *function-name*(*f*) = *axiom-name*(*ax*)
      ∧ $v_{BOUND}$ = (*axiom-argument-list*(*ax*) ∩ V)
      ∧ *derived-from*(BOOLEAN, *function-value-type*(*f*))
      ∧ *legal-condition*(*axiom-antecedent*(*f*), θ, $v_{BOUND}$, *d*)
      ∧ ∀*v* ∈ *function-arguments*(*f*):
        *assignable-to*(*function-argument-type*(*f*, *v*), *axiom-argument*(*ax*, *v*), θ, *d*)

   *legal-action*(*a*, *d*) ≡
     ∃θ, $v_{BOUND}$:
      ∀*v* ∈ *action-parameters*(*a*): (*assignable-to*(*action-parameter-type*(*a*, *v*), *v*, θ, *d*)
                  ∧ *action-parameter-type*(*a*, *v*) ∈ *domain-types*(*d*))
      ∧ *assignable-to*(AGENT, *action-performer*(*a*), θ, *d*)
      ∧ $v_{BOUND}$ = *bound-variables-in-conditions*(*cnd*, $v_{BOUND}$)
              ∪ *action-parameters*(*a*) ∪ ({*action-performer*(*a*)} ∩ V)
      ∧ ∀*cnd* ∈ *action-preconditions*(*a*): *legal-condition*(*cnd*, θ, $v_{BOUND}$, *d*)
      ∧ ∀*e* ∈ *action-effects*(*a*): *legal-effect*(*e*, θ, $v_{BOUND}$, *d*)

*legal-event*(*ev*, *d*) ≡
  ∃θ, *v*$_{BOUND}$:
    ∀*v* ∈ *event-qualities*(*ev*): (*assignable-to*(*event-quality-type*(*ev*, *v*), *v*, θ, *d*)
                    ∧ *event-quality-type*(*ev*, *v*) ∈ *domain-types*(*d*))
    ∧ *v*$_{BOUND}$ = *bound-variables-in-conditions*(*cnd*, *v*$_{BOUND}$) ∪ *event-qualities*(*ev*)
    ∧ ∀*cnd* ∈ *event-triggers*(*ev*): *legal-condition*(*cnd*, θ, *v*$_{BOUND}$, *d*)
    ∧ ∀*e* ∈ *event-effects*(*ev*): *legal-effect*(*e*, θ, *v*$_{BOUND}$, *d*)

*legal-process*(*p*, *d*) ≡
  ∃θ, *v*$_{BOUND}$: ∀*v* ∈ *process-qualities*(*p*):
      (*assignable-to*(*process-quality-type*(*p*, *v*), *v*, θ, *d*)
       ∧ *process-quality-type*(*p*, *v*) ∈ *domain-types*(*d*)
    ∧ *v*$_{BOUND}$ = *bound-variables-in-conditions*(*cnd*, *v*$_{BOUND}$) ∪ *process-qualities*(*p*)
    ∧ ∀*cnd* ∈ *process-conditions*(*p*): *legal-condition*(*cnd*, θ, *v*$_{BOUND}$, *d*)
    ∧ ∀*c* ∈ *process-changes*(*p*): *legal-change*(*c*, θ, *v*$_{BOUND}$, *d*)

*legal-domain*(*d*) ≡
  ∀⟨*child*, *parent*⟩ ∈ *domain-supertypes*(*d*): *child* ∈ *domain-types*(*d*) ∧ *parent* ∈ *domain-types*(*d*)
  ∀⟨*object*, *type*⟩ ∈ *domain-constants*(*d*): *type* ∈ *domain-types*(*d*)
  ∧ ∀*f* ∈ *domain-functions*(*d*): *legal-function*(*f*, *d*)
  ∧ ∀*a* ∈ *domain-actions*(*d*): *legal-action*(*a*, *d*)
  ∧ ∀*ev* ∈ *domain-events*(*d*): *legal-event*(*ev*, *d*)
  ∧ ∀*p* ∈ *domain-processes*(*d*): *legal-process*(*p*, *d*)
  ∧ ∀*ax* ∈ *domain-axioms*(*d*): *legal-axiom*(*ax*, *d*)

*legal-ground-fluent*(*gf* = ⟨*fn*, *va*$_{ARGS}$, *va*$_{FV}$⟩ ∈ *Fn* × *Va* × *Va*, *d*) ≡
  ∃*f* ∈ *domain-functions*(*d*):
    *function-name*(*f*) = *fn*
    ∧ |*function-arguments*(*f*)| = |*va*$_{ARGS}$|
    ∧∀*i*, 0 ≤ *i* < |*va*$_{ARGS}$|: *assignable-to*(*function-argument-type*(*f*, *function-arguments*(*f*)$_i$), *va*$_{ARGS, i}$, ∅, *d*)
        ∧ *assignable-to*(*function-value-type*(*f*), *va*$_{FV}$, ∅, *d*)*legal-environment*(*d*, *sg*) ≡
  *legal-domain*(*d*)
  ∧ ∀*gf* ∈ *scenario-generator-fluents*(*sg*):
  ∧ ∀⟨*fn*, *val*⟩ ∈ *scenario-generator-defaults*(*sg*): ∃*f* ∈ *domain-functions*(*d*): *function-name*(*f*) = *fn*
                    ∧ *assignable-to*(*val*, *function-value-type*(*f*), ∅, *d*)
  ∧ ∀*f* ∈ *domain-functions*(*d*): ∃*val*: ⟨*function-name*(*f*), *val*⟩ ∈ *scenario-generator-defaults*(*sg*)
  ∧ ∀*og* ∈ *scenario-generator-object-generators*(*g*): *derived-from*(*object-generator-type(og)*, O$_{BJECT}$, *d*)
  ∧ ∀*fg* ∈ *scenario-generator-fluent-generators*(*sg*):
      ∃*f*: *f* ∈ *domain-functions*(*d*) ∧ *function-name*(*f*) = *fluent-generator-function-name*(*fg*)
      ∧ G(*legal-ground-fluent*(*fluent-generator-draw*(*fg*), ∅, ∅, *d*))
  ∧ *legal-calculation*(R$_{EAL}$, *scenario-generator-performance-calculation*(*sg*), {⟨A$_G$, A$_{GENT}$⟩}, *d*)

# Appendix B   Definition of *apply* function

We can now define the apply function for each of a set of transformations.

*apply*(ADDTYPE (TYPE: *new-type*),
  ⟨*types, supertypes, constantTypes, functions, axioms, actions, events, processes*⟩, *sg*) ≡
 ⟨*types* ∪ {*new-type*}, *supertypes*,
  *constantTypes, functions, axioms, actions, events, processes, perf*⟩

*apply*(ADDTYPEPARENT(CHILD: *child,* PARENT: *parent*),
  ⟨*types, supertypes, constantTypes, functions, axioms, actions, events, processes*⟩, *sg*) ≡
 ⟨*types, supertypes* ∪ {⟨*child, parent*⟩},
  *constantTypes, functions, axioms, actions, events, processes, perf*⟩

*apply*(REMOVETYPEPARENT(CHILD: *child,* PARENT: *parent*),
  ⟨*types, supertypes, constantTypes, functions, axioms, actions, events, processes*⟩, *sg*) ≡
 ⟨*types, supertypes* \ {⟨*child, parent*⟩},
  *constantTypes, functions, actions, events, processes, perf*⟩

*apply*(ADDCONSTANT(NAME: *name,* TYPE: *type*),
  ⟨*types, supertypes, constantTypes, functions, axioms, actions, events, processes*⟩, *sg*) ≡
 ⟨*types, supertypes, constantTypes* ∪ {⟨*name, type*⟩},
  *functions, axioms, actions, events, processes, perf*⟩

*apply*(ADDFUNCTION(FUNCTION: *new-function*),
  ⟨*types, supertypes, constantTypes, functions, axioms, actions, events, processes*⟩, *sg*) ≡
 ⟨*types, supertypes, constantTypes, functions* ∪ {*new-function*},
  *axioms, actions, events, processes, perf*⟩

*apply*(ADDAXIOM(AXIOM: *axiom*),
  ⟨*types, supertypes, constantTypes, functions, axioms, actions, events, processes*⟩, *sg*) ≡
 ⟨*types, supertypes, constantTypes, functions,*
  *axioms* ∪ {*axiom*}, *actions, events, processes, perf*⟩

*apply*(REMOVETYPE(NAME: *name*),
  ⟨*types, supertypes, constantTypes, functions, axioms, actions, events, processes*⟩, *sg*) ≡
 ⟨*types* \ *name,*
  *supertypes, constantTypes, functions, axioms, actions, events, processes, perf*⟩

*apply*(REMOVEFUNCTION(NAME: *name*),
    ⟨*types, supertypes, constantTypes, functions, axioms, actions, events, processes*⟩, *sg*) ≡
    ⟨*types, supertypes, constantTypes,*
        {*f* ∈ *functions* | *function-name*(*f*) ≠ *name*}, *axioms, actions, events, processes, perf*⟩

*apply*(REMOVEAXIOM(AXIOM: *axiom*),
    ⟨*types, supertypes, constantTypes, functions, axioms, actions, events, processes*⟩, *sg*) ≡
    ⟨*types, supertypes, constantTypes, functions,*
        *axioms* \ {*axiom*}, *actions, events, processes, perf*⟩

*apply*(ADDACTION(ACTIONNAME: *name*, PERFORMER: *performer*, PARAMETERS: *params*,
            PARAMETERTYPES: *paramTypes*),
    ⟨*types, supertypes, constantTypes, functions, axioms, actions, events, processes*⟩, *sg*) ≡
    ⟨*types, supertypes, constantTypes, functions, axioms,*
        *actions* ∪ {⟨*name, performer, args, argTypes,* [], []⟩}, *events, processes, perf*⟩

*apply*(REMOVEACTION(ACTIONNAME: *name*),
    ⟨*types, supertypes, constantTypes, functions, axioms, actions, events, processes*⟩, *sg*) ≡
    ⟨*types, supertypes, constantTypes, functions, axioms,*
        {*a* ∈ *actions* | *action-name*(*a*) ≠ *name*}, *events, processes, perf*⟩

*apply*(*t* = ADDPRECONDITION(ACTIONNAME: *name*, PRECONDITION: *new-prec*),
    ⟨*types, supertypes, constantTypes, functions, axioms, actions, events, processes*⟩, *sg*) ≡
    ⟨*types, supertypes, constantTypes, functions, axioms,*
        {*applyToAction*(*t, a*) | *a* ∈ *actions*}, *events, processes, perf*⟩

*applyToAction*(ADDPRECONDITION(ACTIONNAME: *name*, PRECONDITION: *new-prec*),
        ⟨*name, perf, args, argTypes, precs, effs*⟩) ≡
    ⟨*name, perf, args, argTypes, precs* ∪ {*new-prec*}, *effs*⟩

*applyToAction*(ADDPRECONDITION(ACTIONNAME: *name*, PRECONDITION: *new-prec*),
        *a* = ⟨*n* ≠ *name, perf, args, argTypes, precs, effs*⟩) ≡ *a*

*apply*(*t* = REMOVEPRECONDITION(ACTIONNAME: *name*, PRECONDITION: *old-prec*),
    ⟨*types, supertypes, constantTypes, functions, axioms, actions, events, processes*⟩, *sg*) ≡
    ⟨*types, supertypes, constantTypes, functions, axioms,* {*applyToAction*(*t, a*) | *a* ∈ *actions*},
        *events, processes, perf*⟩

*applyToAction*(REMOVEPRECONDITION(ACTIONNAME: *name*, PRECONDITION: *old-prec*),
        ⟨*name, args, perf, precs, effs*⟩) ≡
    ⟨ *name, perf, args, argTypes, precs* \ {*old-prec*}, *effs*⟩

*applyToAction*(R<small>EMOVE</small>P<small>RECONDITION</small>(A<small>CTION</small>N<small>AME</small>: *name*, P<small>RECONDITION</small>: *old-prec*),
  *a* = ⟨*n ≠ name, perf, args, argTypes, precs, effs*⟩) ≡ *a*

*apply*(*t* = A<small>DD</small>A<small>CTION</small>E<small>FFECT</small>(A<small>CTION</small>N<small>AME</small>: *name*, E<small>FFECT</small>: *new-effect*),
  ⟨*types, supertypes, constantTypes, functions, axioms, actions, events, processes*⟩, *sg*) ≡
  ⟨*types, supertypes, constantTypes, functions, axioms,*
    {*applyToAction*(*t, a*) | *a* ∈ *actions*}, *events, processes, perf*⟩

*applyToAction*(A<small>DD</small>A<small>CTION</small>E<small>FFECT</small>(A<small>CTION</small>N<small>AME</small>: *name*, E<small>FFECT</small>: *new-effect*),
  ⟨*name, perf, args, argTypes, precs, effs*⟩) ≡
  ⟨ *name, perf, args, argTypes, precs, effs* ∪ {*new-effect*}⟩

*applyToAction*(A<small>DD</small>A<small>CTION</small>E<small>FFECT</small>(A<small>CTION</small>N<small>AME</small>: *name*, E<small>FFECT</small>: *new-effect*),
  *a* = ⟨*n ≠ name, perf, args, argTypes, precs, effs*⟩) ≡ *a*

*apply*(*t* = R<small>EMOVE</small>A<small>CTION</small>E<small>FFECT</small>(A<small>CTION</small>N<small>AME</small>: *name*, E<small>FFECT</small>: *old-effect*),
  ⟨*types, supertypes, constantTypes, functions, axioms, actions, events, processes*⟩, *sg*) ≡
  ⟨*types, supertypes, constantTypes, functions, axioms,*
    {*applyToAction*(*t, a*) | *a* ∈ *actions*}, *ev* ∈ *events, processes, perf*⟩

*applyToAction*(R<small>EMOVE</small>A<small>CTION</small>E<small>FFECT</small>(A<small>CTION</small>N<small>AME</small>: *name*, E<small>FFECT</small>: *old-effect*),
  ⟨*name, perf, args, argTypes, precs, effs*⟩) ≡
  ⟨*name, perf, args, argTypes, precs, effs* \ {*old-effect*}⟩

*applyToAction*(R<small>EMOVE</small>A<small>CTION</small>E<small>FFECT</small>(A<small>CTION</small>N<small>AME</small>: *name*, E<small>FFECT</small>: *old-effect*),
  *a* = ⟨*n ≠ name, perf, args, argTypes, precs, effs*⟩) ≡ *a*

*apply*(*t* = A<small>DD</small>E<small>VENT</small>(E<small>VENT</small>N<small>AME</small>: *name*, Q<small>UALITIES</small>: *quals,* Q<small>UALITY</small>T<small>YPES</small>: *qualTypes*),
  ⟨*types, supertypes, constantTypes, functions, axioms, actions, events, processes*⟩, *sg*) ≡
  ⟨*types, supertypes, constantTypes, functions, axioms, actions,*
    *events* ∪ {⟨*name, quals, qualTypes,* 1, 0, ∅, ∅⟩}, *processes, perf*⟩

*apply*(R<small>EMOVE</small>E<small>VENT</small>(N<small>AME</small>: *name*),
  ⟨*types, supertypes, constantTypes, functions, axioms, actions, events, processes*⟩, *sg*) ≡
  ⟨*types, supertypes, constantTypes, functions, axioms, actions,*
    {*ev* ∈ *events* | *event-name*(*ev*) ≠ *name*}, *processes, perf*⟩

*apply*(*t* = C<small>HANGE</small>F<small>REQUENCY</small>(E<small>VENT</small>N<small>AME</small>: *name*, F<small>REQUENCY</small>: *new-freq*),
  ⟨*types, supertypes, constantTypes, functions, axioms, actions, events, processes*⟩, *sg*) ≡
  ⟨*types, supertypes, constantTypes, functions, axioms, actions,*
    {*applyToEvent*(*t, ev*) | *ev* ∈ *events*}, *processes* ∪ {*new-process*}⟩

*applyToEvent*(CHANGEFREQUENCY(EVENTNAME: *name*, FREQUENCY: *new-freq*),
    ⟨*name, quals, qualTypes, prob, freq, triggers, effs*⟩) ≡
  ⟨*name, quals, qualTypes, prob, new-freq, triggers, effs*⟩

*applyToEvent*(CHANGEFREQUENCY(EVENTNAME: *name*, FREQUENCY: *new-freq*),
    *ev* = ⟨*n ≠ name, quals, qualTypes, prob, freq, triggers, effs*⟩) ≡ *ev*

*apply*(*t* = CHANGEPROBABILITY(EVENTNAME: *name*, PROBABILITY: *new-prob*),
    ⟨*types, supertypes, constantTypes, functions, axioms, actions, events, processes*⟩, *sg*) ≡
  ⟨*types, supertypes, constantTypes, functions, axioms, actions,*
    {*applyToEvent*(*t, ev*) | *ev* ∈ *events*}, *processes*⟩

*applyToEvent*(CHANGEPROBABILITY(EVENTNAME: *name*, PROBABILITY: *new-prob*),
    ⟨*name, quals, qualTypes, prob, freq, triggers, effs*⟩) ≡
  ⟨*name, quals, qualTypes, new-prob, freq, triggers, effs*⟩

*applyToEvent*(CHANGEPROBABILITY(EVENTNAME: *name*, PROBABILITY: *new-prob*),
    *ev* = ⟨*n ≠ name, quals, qualTypes, prob, freq, triggers, effs*⟩) ≡ *ev*

*apply*(*t* = ADDTRIGGER(EVENTNAME: *name*, PRECONDITION: *new-prec*),
    ⟨*types, supertypes, constantTypes, functions, axioms, actions, events, processes*⟩, *sg*) ≡
  ⟨*types, supertypes, constantTypes, functions, axioms,*
    *actions,* {*applyToEvent*(*t, ev*) | *ev* ∈ *events*}, *processes, perf*⟩

*applyToEvent*(ADDTRIGGER(EVENTNAME: *name*, PRECONDITION: *new-prec*),
    ⟨*name, quals, qualTypes, prob, freq, triggers, effs*⟩) ≡
  ⟨*name, quals, qualTypes, prob, precs* ∪ {*new-prec*}, *effs*⟩

*applyToEvent*(ADDTRIGGER(EVENTNAME: *name*, PRECONDITION: *new-prec*),
    *ev* = ⟨*n ≠ name, quals, qualTypes, prob, freq, triggers, effs*⟩) ≡ *ev*

*apply*(*t* = REMOVETRIGGER(EVENTNAME: *name*, TRIGGER: *old-trigger*),
    ⟨*types, supertypes, constantTypes, functions, axioms, actions, events, processes*⟩, *sg*) ≡
  ⟨*types, supertypes, constantTypes, functions, axioms, actions,*
    {*applyToEvent*(*t, ev*) | *ev* ∈ *events*}, *processes, perf*⟩

*applyToEvent*(REMOVETRIGGER(EVENTNAME: *name*, TRIGGER: *old-trigger*),
    ⟨*name, quals, qualTypes, prob, triggers, effs*⟩) ≡
  ⟨*name, quals, qualTypes, prob, freq, triggers* \ {*old-trigger*}, *effs*⟩

*applyToEvent*(REMOVETRIGGER(EVENTNAME: *name*, TRIGGER: *old-trigger*),
    *ev* = ⟨*n ≠ name, quals, qualTypes, prob, freq, triggers, effs*⟩) ≡ *ev*

*apply*(*t* = AddEventEffect(eventName: *name*, effect: *new-effect*),
⟨*types, supertypes, constantTypes, functions, axioms, actions, events, processes*⟩, *sg*) ≡
⟨*types, supertypes, constantTypes, functions, axioms,*
*actions,* {*applyToEvent*(*t, ev*) | *ev* ∈ *events*}, *processes, perf*⟩

*applyToEvent*(AddEventEffect(eventName: *name*, effect: *new-effect*),
⟨*name, quals, qualTypes, prob, freq, triggers, effs*⟩) ≡
⟨*name, quals, qualTypes, prob, freq, triggers, effs* ∪ {*new-effect*}⟩

*applyToEvent*(AddEventEffect(eventName: *name*, effect: *new-effect*),
*ev* = ⟨*n* ≠ *name, quals, qualTypes, prob, freq, triggers, effs*⟩) ≡ *ev*

*apply*(*t* = RemoveEventEffect(eventName: *name*, effect: *old-effect*),
⟨*types, supertypes, constantTypes, functions, axioms, actions, events, processes*⟩, *sg*) ≡
⟨*types, supertypes, constantTypes, functions, axioms,*
*actions,* {*applyToEvent*(*t, ev*) | *ev* ∈ *events*}, *processes, perf*⟩

*applyToEvent*(RemoveEventEffect(eventName: *name*, effect: *old-effect*),
⟨*name, quals, qualTypes, prob, freq, triggers, effs*⟩) ≡
⟨*name, quals, qualTypes, prob, freq, triggers, effs* \ {*old-effect*}⟩

*applyToEvent*(RemoveEventEffect(eventName: *name*, effect: *old-effect*),
*ev* = ⟨*n* ≠ *name, quals, qualTypes, prob, freq, triggers, effs*⟩) ≡ *ev*

*apply*(AddProcess(processName: *name,* qualities: *quals,* qualityTypes: *qualTypes*),
⟨*types, supertypes, constantTypes, functions, axioms, actions, events, processes*⟩, *sg*) ≡
⟨*types, supertypes, constantTypes, functions, axioms, actions, events,*
*processes* ∪ {⟨*name, quals, qualTypes,* ∅, ∅⟩}, *perf*⟩

*apply*(RemoveProcess(name: *name*),
⟨*types, supertypes, constantTypes, functions, axioms, actions, events, processes*⟩, *sg*) ≡
⟨*types, supertypes, constantTypes, functions, axioms, actions, events,*
{*p* ∈ *processes* | *process-name*(*p*) ≠ *name*}⟩

*apply*(*t* = AddProcessCondition(processName: *name*, condition: *new-condition*),
⟨*types, supertypes, constantTypes, functions, axioms, actions, events, processes*⟩, *sg*) ≡
⟨*types, supertypes, constantTypes, functions, axioms, actions, events,*
{*applyToProcess*(*t, p*) | *p* ∈ *processes*}⟩

*applyToProcess*(AddProcessCondition(processName: *name*, condition: *new-condition*),
⟨*name, quals, qualTypes, conds, changes*⟩) ≡
⟨*name, quals, qualTypes, conds* ∪ {*new-condition*}, *changes*⟩

*applyToProcess*(AddProcessCondition(processName: *name*, condition: *new-condition*),
*p* = ⟨*n* ≠ *name, quals, qualTypes, conds, changes*⟩) ≡ *p*

*apply*(*t* = REMOVEPROCESSCONDITION(PROCESSNAME: *name*, CONDITION: *old-condition*),
⟨*types, supertypes, constantTypes, functions, axioms, actions, events, processes*⟩, *sg*) ≡
⟨*types, supertypes, constantTypes, functions, axioms, actions, events,*
{*applyToProcess*(*t, p*) | *p* ∈ *processes*}⟩

*applyToProcess*(REMOVEPROCESSCONDITION(PROCESSNAME: *name*, CONDITION: *old-condition*),
⟨*name, quals, qualTypes, conds, changes*⟩) ≡
⟨*name, quals, qualTypes, conds* \ {*old-condition*}, *changes*⟩

*applyToProcess*(REMOVEPROCESSCONDITION(PROCESSNAME: *name*, CONDITION: *old-condition*),
*p* = ⟨*n* ≠ *name, quals, qualTypes, conds, changes*⟩) ≡ *p*

*apply*(*t* = ADDPROCESSCHANGE(PROCESSNAME: *name*, CHANGE: *new-change*),
⟨*types, supertypes, constantTypes, functions, actions, events, processes*⟩, *sg*) ≡
⟨*types, supertypes, constantTypes, functions, axioms, actions, events,*
{*applyToProcess*(*t, p*) | *p* ∈ *processes*}⟩

*applyToProcess*(ADDPROCESSCHANGE(PROCESSNAME: *name*, CHANGE: *new-change*),
⟨*name, quals, qualTypes, conds, changes*⟩) ≡
⟨*name, quals, qualTypes, conds, changes* \ {*new-change*}⟩

*applyToProcess*(ADDPROCESSCHANGE(PROCESSNAME: *name*, CHANGE: *new-change*),
*p* = ⟨*n* ≠ *name, quals, qualTypes, conds, changes*⟩) ≡ *p*

*apply*(*t* = REMOVEPROCESSCHANGE(PROCESSNAME: *name*, CHANGE: *old-change*),
⟨*types, supertypes, constantTypes, functions, axioms, actions, events, processes*⟩, *sg*) ≡
⟨*types, supertypes, constantTypes, functions, axioms, actions, events,*
{*applyToProcess*(*t, p*) | *p* ∈ *processes*}⟩

*applyToProcess*(REMOVEPROCESSCHANGE(PROCESSNAME: *name*, CHANGE: *old-change*),
⟨*name, quals, qualTypes, conds, changes*⟩) ≡
⟨*name, quals, qualTypes, conds, changes* \ {*old-change*}⟩

*applyToProcess*(REMOVEPROCESSCHANGE(PROCESSNAME: *name*, CHANGE: *old-change*),
*p* = ⟨*n* ≠ *name, quals, qualTypes, conds, changes*⟩) ≡ *p*

# Appendix C  Differences between T-SAL Domain Language & PDDL+

T-SAL-CR is closely related to PDDL+, due to its use of events and processes to model mixed discrete-continuous domains in a general way. While there is not yet a complete semantics for simulating T-SAL, we expect that one will be provided later. The following differences between T-SAL and PDDL+ are relevant to understanding the language:

- CREATE effects: Actions and events can result in the creation of new objects. Open worlds need to allow for new objects to arise as part of the dynamics.
- Object fluents: T-SAL makes use of object fluents, not just numeric fluents. Object fluents make many expressions more succinct and permit the description of one-to-one relationships between objects.
- Calculations: These are comparable to PDDL+ fluent expressions; T-SAL adds mathematical aggregation over a class of objects, a trinary "if" operator, and recursion through function terms.
- Conditionals: These are comparable to PDDL+ goal descriptions; T-SAL adds recursion through function terms.
- Process changes: T-SAL allows arbitrary time differential equations in process changes, whereas PDDL+ incorporates only linear differentials. T-SAL uses DT instead of #T to signify the differential operator, which is cosmetic.
- Probabilistic events: Whereas PDDL+ events occur deterministically, T-SAL events occur nondetermistically, either according to a probability or at times drawn from a distribution.
- Action performers: To describe multi-agent domains, T-SAL adds a performer property to agent models that allows preconditions that refer to the agent performing an action.
- Name changes: To reflect the fact that event and process variables are outside of agent control, we rename "parameters" to "qualities". To reflect the fact that the occurrence of an event is required, we use the term "triggers" instead of "preconditions". Processes have "conditions" instead of "preconditions" or "triggers" as they must be active for an indefinite period.
- Conditional and probabilistic effects: Not present in T-SAL, to avoid redundancies with events.
- "Exists" goals: Not present in T-SAL, as all free variables are implicitly existentially quantified.
- "Forall" effects: Not present in T-SAL, events are intended to handle changes to all objects in a class.
- Durative actions: Not present in T-SAL, these are redundant with processes and events.
- Predicates: Not present in T-SAL, redundant with fluents.
- "Either" types: Not present in T-SAL.

# Appendix D  T-SAL Computational Representation Backus-Naur Form

| | |
|---|---|
| <DOMAIN> | ::= (DEFINE (DOMAIN <NAME>) |
| | [<TYPES-DEF>] |
| | [<SUPERTYPES-DEF>] |
| | [<CONSTANTS-DEF>] |
| | [<FUNCTIONS-DEF>] |
| | [<STRUCTURE-DEF>*] |
| <TYPES-DEF> | ::= (:TYPES <TYPED LIST (TYPE)>*) |
| <CONSTANTS-DEF> | ::= (:CONSTANTS <TYPED LIST (OBJECT)>) |
| <FUNCTIONS-DEF> | ::= (:FUNCTIONS <TYPED LIST (FUNCTION-DEF)>) |
| <FUNCTION-DEF> | ::= (<FUNCTION-NAME> <TYPED LIST (VARIABLE)>*) |
| <STRUCTURE-DEF> | ::= <ACTION-MODEL-DEF> |
| <STRUCTURE-DEF> | ::= <EVENT-MODEL-DEF> |
| <STRUCTURE-DEF> | ::= <PROCESS-MODEL-DEF> |
| <STRUCTURE-DEF> | ::= <AXIOM-DEF> |
| <ACTION-MODEL-DEF> | ::= (:ACTION <NAME> |
| | :PERFORMER <VARIABLE> |
| | :PARAMETERS (<TYPED-LIST (VARIABLE)>) |
| | :PRECONDITIONS (<CONDITION>*) |
| | :EFFECTS (<EFFECT-DEF>*)) |
| <EVENT-MODEL-DEF> | ::= (:EVENT <NAME> |
| | [:PROBABILITY <REAL>] |
| | [:FREQUENCY <REAL>] |
| | :QUALITIES (<TYPED-LIST (VARIABLE)>) |
| | :TRIGGERS (<CONDITION>*) |
| | :EFFECTS (<EFFECT-DEF>*)) |
| <PROCESS-MODEL-DEF> | ::= (:PROCESS <NAME> |
| | :QUALITIES (<TYPED-LIST (VARIABLE)>) |
| | :CONDITIONS (<CONDITION>*) |
| | :CHANGES (<CHANGE-DEF>*)) |
| <AXIOM-DEF> | ::= (:- <FUNCTION-CALL> <CONDITION>) |
| <CONDITION> | ::= <TERM> |
| <CONDITION> | ::= (<INEQUALITY> <TERM> <TERM>) |
| <CONDITION> | ::= (OR\|AND <CONDITION> <CONDITION>+) |
| <CONDITION> | ::= (NOT <CONDITION>) |
| <CONDITION> | ::= (FORALL (VARIABLE+) <CONDITION> <CONDITION>) |
| <CALCULATION> | ::= <TERM> |
| <CALCULATION> | ::= <FLUENT-REF> |
| <CALCULATION> | ::= (<MOP> <CALCULATION> <CALCULATION>) |
| <CALCULATION> | ::= (SUM \| PRODUCT <VARIABLE> <CONDITION> <CALCULATION>) |
| <CALCULATION> | ::= (IF <CONDITION> <CALCULATION> <CALCULATION>) |
| <EFFECT-DEF> | ::= (<MOD> (<FUNCTION-NAME> <BASIC-TERM>*) <TERM>) [<PROBABILITY>] |
| <CHANGE-DEF> | ::= (<DIR> (<FUNCTION-NAME> <BASIC-TERM>*) <CALCULATION>) |
| <TERM> | ::= <BASIC-TERM> |
| <TERM> | ::= <VARIABLE> |
| <TERM> | ::= (<FUNCTION-NAME> <TERM>*) |
| <BASIC-TERM> | ::= <OBJECT> |
| <BASIC-TERM> | ::= <NUMBER> |
| <BASIC-TERM> | ::= TRUE \| FALSE |
| <MOP> | ::= + \| - \| * \| / \| :UNIFORM \| :GAUSSIAN |
| <MOD> | ::= SET \| INCREASE \| DECREASE \| CREATE |
| <DIR> | ::= INCREASE \| DECREASE |
| <TYPED LIST ($x$)> | ::= $x$* |
| <TYPED LIST ($x$)> | ::= $x^+$ - <TYPE> <TYPED LIST ($x$)>* |
| <TYPE> | ::= <NAME> |
| <FUNCTION-NAME> | ::= <NAME> |
| <OBJECT> | ::= <NAME> |
| <VARIABLE> | ::= ?<NAME> |
| <INEQUALITY> | ::= "=" \| "!=" \| "<" \| ">" \| "<=" \| ">=" |

Table 4. Backus-Naur Form specification of T-SAL-CR